\documentclass[conference]{IEEEtran}
\IEEEoverridecommandlockouts
\usepackage[letterpaper,left=56pt,right=56pt,top=54pt,bottom=57pt]{geometry}
\def\IEEEtitletopspaceextra{18pt}
\usepackage{cite}
\usepackage{amsmath,amssymb,amsfonts}
\usepackage{graphicx}
\usepackage{textcomp}
\usepackage{xcolor}
\usepackage{booktabs}
\usepackage{multirow}
\usepackage[protrusion=false]{microtype}
\usepackage{array}
\usepackage{adjustbox}

\usepackage{capt-of}
\usepackage{siunitx}
\usepackage{url}
\usepackage{placeins}
\usepackage[hidelinks]{hyperref}

\newif\ifanon \anonfalse

\definecolor{tbdred}{rgb}{1,0,0}\newcommand{\tbd}[1]{\textcolor{tbdred}{\textbf{[TBD:} #1\textbf{]}}}
\newcommand{\figtbd}[1]{\begin{center}\textcolor{red}{\textbf{[FIGURE TBD:} #1\textbf{]}}\end{center}}
\newcommand{\phfig}[2]{\fbox{\parbox[c][#1][c]{\dimexpr\linewidth-2\fboxsep-2\fboxrule\relax}{\centering\footnotesize\textcolor{red}{\textbf{[FIGURE TBD:} #2\textbf{]}}}}}
\IfFileExists{macros.tex}{\newcommand{\LidarInverted}{inverted}

\newcommand{\LidarHeightM}{2.46}
\newcommand{\CamDropM}{0.58}

\newcommand{\FovSumDeg}{423.6}

\newcommand{\InsModel}{SBG Ellipse-D}

\newcommand{\SyncOffsetUs}{26}
\newcommand{\SyncStdUs}{\textcolor{red}{196}}

\newcommand{\LatencyResMs}{3}
\newcommand{\IntrAcceptPx}{\textcolor{red}{0.5}}
\newcommand{\IntrAcceptCoverage}{\textcolor{red}{50}}

\newcommand{\RecGbPerHour}{\textcolor{red}{689}}
\newcommand{\RecGbPerHourCompressed}{\textcolor{red}{301}}
\newcommand{\ChunkSeconds}{300}

\newcommand{\GnssGateM}{\textcolor{red}{0.30}}

\newcommand{\NClasses}{8}

\newcommand{\DensityRadiusM}{30}
\newcommand{\DensityQuantiles}{25/50/75}

\newcommand{\RouteSepM}{100}
\newcommand{\SplitFractions}{\textcolor{red}{70/15/15}}

\newcommand{\ArmBTrainImages}{\textcolor{red}{18,681}}
\newcommand{\PilotKeyframes}{\textcolor{red}{1,364}}

\newcommand{\NArmA}{\textcolor{red}{66,542}}
\newcommand{\NArmB}{\textcolor{red}{6,136}}

\newcommand{\NMerged}{\textcolor{red}{90,667}}
\newcommand{\NSuppressed}{\textcolor{red}{2,651}}
\newcommand{\NVetoed}{\textcolor{red}{408}}

\newcommand{\NMasksKept}{\textcolor{red}{87,063}}

\newcommand{\NPainted}{\textcolor{red}{14.4\,M}}
\newcommand{\NBoxes}{\textcolor{red}{74,369}}

\newcommand{\NReachable}{\textcolor{red}{8}}

\newcommand{\NMarkers}{\textcolor{red}{12}}
\newcommand{\NDegraded}{\textcolor{red}{7}}

\newcommand{\SessNScenes}{\textcolor{red}{11}}

\newcommand{\SessNCameras}{6}
\newcommand{\SessMinutes}{\textcolor{red}{50}}
\newcommand{\SessKeyframeHz}{5}

\newcommand{\RelNChunksDone}{5}

\newcommand{\DoubleFractionPct}{5}
\newcommand{\IllumEdges}{\textcolor{red}{45, 75, 95}}
\newcommand{\StitchMaxGap}{3}
\newcommand{\StitchBaseGateM}{\textcolor{red}{2.0}}
\newcommand{\StitchSlackM}{\textcolor{red}{1.0}}
\newcommand{\StitchSizeRatio}{\textcolor{red}{2.0}}
\newcommand{\StitchMinPts}{\textcolor{red}{5}}
\newcommand{\MovingSpeedMps}{0.5}
\newcommand{\AttrMaxDtS}{\textcolor{red}{1.5}}
\newcommand{\RelNSamples}{\textcolor{red}{5,996}}
\newcommand{\RelNAnnotations}{109,813}
\newcommand{\RelNAnnotationsK}{109.8k}
\newcommand{\RelNInstances}{\textcolor{red}{29,379}}
\newcommand{\RelNExcluded}{\textcolor{red}{120,297}}

\newcommand{\RelChainHours}{\textcolor{red}{39}}
\newcommand{\RelPropagationHours}{\textcolor{red}{26}}
\newcommand{\RelTracksConfirmed}{\textcolor{red}{426}}

\newcommand{\BenchNClasses}{8}
\newcommand{\BenchNHeadline}{\textcolor{red}{2}}

\newcommand{\BenchConfigHash}{\textcolor{red}{e54c8267837cc82b}}

\newcommand{\GenSpreadBEVFusionOne}{\textcolor{red}{4.0}}
\newcommand{\GenSpreadBEVFusionFive}{\textcolor{red}{1.8}}
\newcommand{\PooledCngHeldOutAP}{\textcolor{red}{62}}

\newcommand{\DayTwoCenterPointMeanmAP}{38.7}

\newcommand{\DayTwoCenterPointMeanRickshawAP}{54.2}

\newcommand{\DayTwoCenterPointMaxSeedSd}{0.7}
\newcommand{\DayTwoBEVFusionMeanmAP}{\textcolor{red}{31.3}}

\newcommand{\DayTwoPointPillarsMeanmAP}{\textcolor{red}{24.1}}

\newcommand{\DayTwoTransFusionMeanmAP}{\textcolor{red}{21.7}}

\newcommand{\DensDeltaCenterPoint}{+6.3}
\newcommand{\RsCenterPoint}{\textcolor{red}{0.322}}

\newcommand{\DensDeltaPointPillars}{\textcolor{red}{+4.1}}
\newcommand{\RsPointPillars}{\textcolor{red}{0.238}}

\newcommand{\StrataGtLow}{960}
\newcommand{\StrataGtExtreme}{4,884}

\newcommand{\OdomKISSICPMinATE}{2.9}
\newcommand{\OdomKISSICPMaxATE}{6.0}

\newcommand{\OdomGLIMMinATE}{2.3}
\newcommand{\OdomGLIMMaxATE}{4.0}

\newcommand{\PpkCorrMinATE}{1.4}
\newcommand{\PpkCorrMaxATE}{6.9}
\newcommand{\DepthAbsRelMean}{0.92}
\newcommand{\DepthRmseMean}{10.4}

\newcommand{\DepthNightAbsRelPenaltyPct}{\textcolor{red}{20}}
\newcommand{\DepthNightRmsePenaltyPct}{\textcolor{red}{13}}

\newcommand{\LocGLIMLat}{1.07}
\newcommand{\LocGLIMRatio}{1.28}

\newcommand{\LocKISSICPLat}{0.83}
\newcommand{\LocKISSICPRatio}{1.66}
\newcommand{\OccVoxelMean}{18.8}
\newcommand{\OccRayMean}{73.1}
\newcommand{\OccRatio}{3.9}
\newcommand{\CloudStereoSharePct}{\textcolor{red}{90.4}}
\newcommand{\CloudMidSharePct}{\textcolor{red}{9.6}}
\newcommand{\KissDayOneMidATE}{\textcolor{red}{3.86}}
\newcommand{\KissDayOneFusedATE}{\textcolor{red}{130.95}}
\newcommand{\KissDayOneStereoATE}{\textcolor{red}{199.05}}
\newcommand{\BoxAnchorCoveragePct}{\textcolor{red}{39}}
\newcommand{\LocalisedPedEightPct}{\textcolor{red}{65}}
\newcommand{\LocalisedPedNinePct}{\textcolor{red}{84}}
\newcommand{\ScoreAboveHalfPct}{\textcolor{red}{0.36}}
\newcommand{\GtPerFrameZero}{\textcolor{red}{30}}
\newcommand{\GtPerFrameEight}{\textcolor{red}{3}}
\newcommand{\GtPerFrameNine}{\textcolor{red}{10}}
\newcommand{\MidPtsPerFrame}{\textcolor{red}{39,936}}
\newcommand{\ZedPtsPerFrame}{\textcolor{red}{166,816}}
\newcommand{\MidPtsPerBoxLo}{\textcolor{red}{15}}
\newcommand{\MidPtsPerBoxHi}{\textcolor{red}{25}}
\newcommand{\MidEmptyBoxLoPct}{\textcolor{red}{9.4}}
\newcommand{\MidEmptyBoxHiPct}{\textcolor{red}{12.6}}
\newcommand{\PpkDayTwoFixedPct}{47.0}
\newcommand{\PpkDayTwoFloatPct}{52.8}
\newcommand{\PpkDayTwoEpochs}{\textcolor{red}{3,977}}
\newcommand{\PpkGpstOffsetS}{\textcolor{red}{18}}
\newcommand{\PpkBoresightDeg}{\textcolor{red}{87.8}}
\newcommand{\PpkBoresightMadDeg}{\textcolor{red}{1.89}}

\newcommand{\DayTwoMaxRangeM}{\textcolor{red}{30.7}}

\newcommand{\DayTwoLinkedPct}{\textcolor{red}{94.3}}

\newcommand{\DayTwoAnchorCoveragePct}{100}

\newcommand{\MidRangeStepCm}{\textcolor{red}{8}}

\newcommand{\ShockPeakAccel}{\textcolor{red}{38.5}}
\newcommand{\ShockPeakG}{3.9}
\newcommand{\TrackMedianFrames}{\textcolor{red}{3}}

\newcommand{\SegLidarLabelledPct}{\textcolor{red}{8.78}}
\newcommand{\SegZedLabelledPct}{\textcolor{red}{8.62}}
\newcommand{\SegBevfusionThreshold}{\textcolor{red}{0.002}}
\newcommand{\OccSweepFillPct}{0.55}
\newcommand{\OccAccumFillPct}{3.30}
\newcommand{\DepthCoincidentPx}{\textcolor{red}{370}}
\newcommand{\DepthGtCoveragePct}{\textcolor{red}{0.4}}

\newcommand{\PpkSigmaFixedMm}{7.3}
\newcommand{\PpkSigmaFloatMm}{76}
\newcommand{\PpkHeadingAgreeDeg}{1.89}
\newcommand{\DropoutLidarS}{15.6}
\newcommand{\DropoutImuS}{\textcolor{red}{5.95}}

\newcommand{\BlackwellHead}{\textcolor{red}{a3d9ce6}}

\newcommand{\NSessionsRecorded}{\textcolor{red}{7}}
\newcommand{\NRecordingDays}{\textcolor{red}{2}}
\newcommand{\RecordedKm}{\textcolor{red}{71}}
\newcommand{\RecordedHours}{\textcolor{red}{5.5}}

}{}
\newcommand{\srcnote}{}\definecolor{red}{rgb}{0,0,0}

\providecommand{\LidarHeightM}{\tbd{h}}\providecommand{\CamDropM}{\tbd{h}}
\providecommand{\FovSumDeg}{\tbd{FOV}}
\providecommand{\LidarInverted}{\tbd{mount}}
\providecommand{\InsModel}{\tbd{INS}}
\providecommand{\GnssGateM}{\tbd{m}}
\providecommand{\SyncOffsetUs}{\tbd{us}}\providecommand{\SyncStdUs}{\tbd{us}}\providecommand{\LatencyResMs}{\tbd{ms}}
\providecommand{\IntrAcceptPx}{\tbd{px}}\providecommand{\IntrAcceptCoverage}{\tbd{\%}}
\providecommand{\RecGbPerHour}{\tbd{GB/h}}\providecommand{\RecGbPerHourCompressed}{\tbd{GB/h}}\providecommand{\ChunkSeconds}{\tbd{s}}
\providecommand{\NClasses}{\tbd{N}}\providecommand{\DensityRadiusM}{\tbd{R}}
\providecommand{\DensityQuantiles}{\tbd{q}}\providecommand{\RouteSepM}{\tbd{m}}\providecommand{\SplitFractions}{\tbd{split}}
\providecommand{\NReachable}{\tbd{N}}\providecommand{\ArmBTrainImages}{\tbd{N}}
\providecommand{\PilotKeyframes}{\tbd{N}}
\providecommand{\NArmA}{\tbd{N}}\providecommand{\NArmB}{\tbd{N}}\providecommand{\NMerged}{\tbd{N}}\providecommand{\NSuppressed}{\tbd{N}}\providecommand{\NVetoed}{\tbd{N}}

\providecommand{\NMasksKept}{\tbd{N}}\providecommand{\NPainted}{\tbd{N}}\providecommand{\NBoxes}{\tbd{N}}
\providecommand{\NMarkers}{\tbd{N}}\providecommand{\NDegraded}{\tbd{N}}

\providecommand{\SessNScenes}{\tbd{N}}\providecommand{\SessNCameras}{\tbd{N}}\providecommand{\SessMinutes}{\tbd{min}}\providecommand{\SessKeyframeHz}{\tbd{Hz}}
\providecommand{\RelNChunksDone}{\tbd{N}}\providecommand{\RelNSamples}{\tbd{N}}\providecommand{\RelNAnnotations}{\tbd{N}}\providecommand{\RelNAnnotationsK}{\tbd{N}}\providecommand{\RelNInstances}{\tbd{N}}\providecommand{\RelNExcluded}{\tbd{N}}

\providecommand{\DoubleFractionPct}{\tbd{\%}}\providecommand{\IllumEdges}{\tbd{edges}}\providecommand{\StitchMaxGap}{\tbd{k}}\providecommand{\StitchBaseGateM}{\tbd{m}}\providecommand{\StitchSlackM}{\tbd{m}}\providecommand{\StitchSizeRatio}{\tbd{r}}\providecommand{\StitchMinPts}{\tbd{n}}\providecommand{\MovingSpeedMps}{\tbd{v}}\providecommand{\AttrMaxDtS}{\tbd{s}}
\newcommand{\DatasetScenes}{175}\newcommand{\DatasetHours}{14\,h 35\,min}\newcommand{\DatasetHoursDecimal}{14.6}
\newcommand{\DatasetSessions}{21}\newcommand{\DatasetKm}{157}\newcommand{\DatasetAreas}{14}
\newcommand{\DatasetRoadPointsM}{225}
\newcommand{\RawLidarPointsB}{7.49}
\newcommand{\RoadShareLidarPct}{9.3}\newcommand{\RoadShareSeenPct}{21.6}\newcommand{\RoadPtsPerKf}{3,700}

\newcommand{\DatasetAdmittedM}{813K}\newcommand{\DatasetWithheldM}{893K}\newcommand{\DatasetFlaggedM}{415K}\newcommand{\DatasetRejectedM}{461K}
\newcommand{\DatasetBoxesM}{2.3M}\newcommand{\DatasetThreeDBoxesM}{813K}
\newcommand{\ProcessedHours}{5\,h}\newcommand{\ProcessedGpuHours}{540}
\newcommand{\DatasetAdmittedPct}{47.7}\newcommand{\DatasetWithheldPct}{52.3}\newcommand{\DatasetFlaggedPct}{24.3}\newcommand{\DatasetRejectedPct}{27.0}
\providecommand{\NSessionsRecorded}{\tbd{N}}\providecommand{\NRecordingDays}{\tbd{N}}\providecommand{\RecordedKm}{\tbd{km}}\providecommand{\RecordedHours}{\tbd{h}}
\providecommand{\RelChainHours}{\tbd{h}}\providecommand{\RelPropagationHours}{\tbd{h}}\providecommand{\RelTracksConfirmed}{\tbd{N}}
\providecommand{\BenchNClasses}{\tbd{N}}\providecommand{\BenchNHeadline}{\tbd{N}}\providecommand{\BenchConfigHash}{\tbd{hash}}

\newcommand{\dsense}{DhakaSense}
\newcommand{\dbench}{\texttt{dbench}}
\newcommand{\rs}{\ensuremath{\mathrm{RS}}}
\newcommand{\dens}{\ensuremath{\rho}}

\title{\fontsize{20}{24}\selectfont \dsense: A Multimodal Driving Dataset and Benchmark for Dense, Heterogeneous Traffic in Dhaka}

\ifanon
  \author{\IEEEauthorblockN{Anonymous ICRA 2027 submission}}
\else
  \author{\IEEEauthorblockN{Mollah Md Saif$^{1}$\quad Hironmoy Roy Rudra$^{1}$\quad Zamiul Rashid$^{1}$\quad Debjoty Bhowmik$^{1}$\\ MD Rafid Khan$^{1}$\quad Md Golam Rabiul Alam$^{1}$\quad Md.~Khalilur Rhaman$^{1}$}\IEEEauthorblockA{\normalsize\upshape $^{1}$BRAC University\\[2pt] \small\ttfamily mollah.mdsaif@bracu.ac.bd, hironmoy.roy.rudra@g.bracu.ac.bd, zamiul.rashid@g.bracu.ac.bd,\\ debjoty.bhowmik@g.bracu.ac.bd, md.rafid.khan@g.bracu.ac.bd,\\ rabiul.alam@bracu.ac.bd, khalilur@bracu.ac.bd}}
\fi

\providecommand{\LidarHeightM}{\tbd{h}}\providecommand{\CamDropM}{\tbd{h}}
\providecommand{\FovSumDeg}{\tbd{FOV}}
\providecommand{\LidarInverted}{\tbd{mount}}
\providecommand{\InsModel}{\tbd{INS}}
\providecommand{\GnssGateM}{\tbd{m}}
\providecommand{\SyncOffsetUs}{\tbd{us}}\providecommand{\SyncStdUs}{\tbd{us}}\providecommand{\LatencyResMs}{\tbd{ms}}
\providecommand{\IntrAcceptPx}{\tbd{px}}\providecommand{\IntrAcceptCoverage}{\tbd{\%}}
\providecommand{\RecGbPerHour}{\tbd{GB/h}}\providecommand{\RecGbPerHourCompressed}{\tbd{GB/h}}\providecommand{\ChunkSeconds}{\tbd{s}}
\providecommand{\NClasses}{\tbd{N}}\providecommand{\DensityRadiusM}{\tbd{R}}
\providecommand{\DensityQuantiles}{\tbd{q}}\providecommand{\RouteSepM}{\tbd{m}}\providecommand{\SplitFractions}{\tbd{split}}
\providecommand{\NReachable}{\tbd{N}}\providecommand{\ArmBTrainImages}{\tbd{N}}
\providecommand{\PilotKeyframes}{\tbd{N}}
\providecommand{\NArmA}{\tbd{N}}\providecommand{\NArmB}{\tbd{N}}\providecommand{\NMerged}{\tbd{N}}\providecommand{\NSuppressed}{\tbd{N}}\providecommand{\NVetoed}{\tbd{N}}

\providecommand{\NMasksKept}{\tbd{N}}\providecommand{\NPainted}{\tbd{N}}\providecommand{\NBoxes}{\tbd{N}}
\providecommand{\NMarkers}{\tbd{N}}\providecommand{\NDegraded}{\tbd{N}}

\providecommand{\SessNScenes}{\tbd{N}}\providecommand{\SessNCameras}{\tbd{N}}\providecommand{\SessMinutes}{\tbd{min}}\providecommand{\SessKeyframeHz}{\tbd{Hz}}
\providecommand{\RelNChunksDone}{\tbd{N}}\providecommand{\RelNSamples}{\tbd{N}}\providecommand{\RelNAnnotations}{\tbd{N}}\providecommand{\RelNInstances}{\tbd{N}}\providecommand{\RelNExcluded}{\tbd{N}}

\providecommand{\DoubleFractionPct}{\tbd{\%}}\providecommand{\IllumEdges}{\tbd{edges}}\providecommand{\StitchMaxGap}{\tbd{k}}\providecommand{\StitchBaseGateM}{\tbd{m}}\providecommand{\StitchSlackM}{\tbd{m}}\providecommand{\StitchSizeRatio}{\tbd{r}}\providecommand{\StitchMinPts}{\tbd{n}}\providecommand{\MovingSpeedMps}{\tbd{v}}\providecommand{\AttrMaxDtS}{\tbd{s}}
\providecommand{\RelChainHours}{\tbd{h}}\providecommand{\RelPropagationHours}{\tbd{h}}\providecommand{\RelTracksConfirmed}{\tbd{N}}
\providecommand{\BenchNClasses}{\tbd{N}}\providecommand{\BenchNHeadline}{\tbd{N}}\providecommand{\BenchConfigHash}{\tbd{hash}}

\begin{document}
\bstctlcite{IEEEexample:BSTcontrol}
\maketitle

\begin{abstract}
Most autonomous driving systems are developed and evaluated on data from structured roads where traffic keeps to marked lanes.
South Asia holds around 25.4\% of the world's population, and its urban traffic often mixes cars, buses, CNG auto-rickshaws, cycle rickshaws, motorcycles and pedestrians on dense, heterogeneous and frequently unmarked roads.
Perception and planning for such traffic can only be trained and tested on 3D data that records where every road user is and how it moves.
There are comparatively few 3D driving datasets that capture this type of traffic and the interactions that occur within it.
We present \dsense, a multimodal driving dataset and benchmark collected in Dhaka, Bangladesh, covering \DatasetHours{} of driving across \DatasetScenes{} scenes from afternoon into dusk.
A low-cost sensor rig combines a Livox Mid-360 LiDAR, stereo depth cameras, six RGB cameras and post-processed GNSS/INS ego pose.
The released sequences include 3D bounding boxes, per-point labels and instance identities across an \BenchNClasses-class taxonomy, with CNG auto-rickshaws and cycle rickshaws represented as distinct categories.
The data are annotated by a semiautomatic pipeline that combines a COCO detector, a regionally fine-tuned detector and foundation-model segmentation lifted into the LiDAR point cloud.
The benchmark covers 3D detection, point segmentation, odometry, localization, depth and occupancy, with every result reported per traffic-density level.
In the initial baselines, detection accuracy varies with the distribution of keyframes across traffic conditions, while LiDAR odometry maintains metre-level error over kilometre-scale sequences.
The dataset, collection rig, annotation pipeline and evaluation protocol are released to support further study of autonomous driving in South Asian urban environments.
\end{abstract}

\section{Introduction}
\label{sec:intro}
In the densely populated cities of South Asia, Southeast Asia and Africa, urban traffic mixes motorised and non-motorised vehicles with pedestrians on roads that are often unmarked, lane discipline is weak, pedestrians cross wherever a gap opens, and traffic rules are bent by drivers and pedestrians alike~\cite{varma2019idd,kumar2025driveindia,baig2024badodd}.
Autonomous-driving perception, however, is benchmarked almost entirely on data from a handful of cities in North America, Europe, East Asia and Singapore~\cite{geiger2012kitti,caesar2020nuscenes,sun2020waymo,wilson2021argoverse2,mao2021once}, where traffic keeps to marked lanes, is sparse by global standards and spans a short list of vehicle types.
Dhaka is among the densest cities on earth. A single block holds cars, buses, CNG auto-rickshaws, battery-driven \emph{easy-bikes}, cycle rickshaws, pushcarts and pedestrians on one unmarked carriageway, and the number of road users around a vehicle can change several-fold within a minute.
No released 3D dataset records this traffic. The datasets that exist for the region are image-only, or their 3D labels include no rickshaw class (Section~\ref{sec:related}).

The purpose of \dsense{} is autonomous navigation in the dense, heterogeneous, unstructured traffic of South Asian cities such as Dhaka.
A vehicle that is to drive through a Dhaka intersection must detect every agent around it, follow each one over time, and anticipate where it will be a few seconds ahead, because right of way there is negotiated by proximity and motion rather than granted by lanes and signals.
Perception and forecasting models learn these behaviours only from data that contains them, and lane-based datasets do not.
We therefore collect synchronized LiDAR, stereo-depth and surround-camera data in dense traffic, label every road user with a 3D box, a class and an identity so that detection, segmentation and, once identities persist, tracking and forecasting can be trained and measured on it, and release the recording and labelling stack so that other cities can do the same.
The release supports detection, segmentation, odometry, localization and occupancy, and the tracks and forecasts the planner will consume are the next step the labels are designed for.

A dataset alone does not settle whether such traffic is harder, or in what regime.
Existing benchmarks stratify results by object properties such as distance, occlusion and LiDAR point count~\cite{geiger2012kitti,sun2020waymo} but not by the state of the scene around the vehicle, and robustness benchmarks measure degradation under synthetic corruptions~\cite{kong2023robo3d}, so the natural \emph{corruption} of dense, heterogeneous traffic deserves the same treatment with real data.

\begin{figure}[!h]
  \centering
  \IfFileExists{fig_teaser.pdf}{\includegraphics[width=\linewidth]{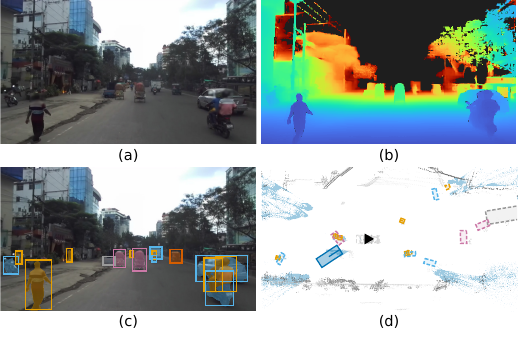}}{\phfig{5.2cm}{run make\_sample\_figs.py}}
  \caption{One \dsense{} keyframe. (a) The front camera frame, (b) the front ZED's 32-bit depth image, reconstructed from the released stereo cloud and coloured by range, (c) the 2D detector boxes and SAM~3.1 instance masks that seed the 3D labels, and (d) the 3D boxes from above, forward to the right, released boxes filled and withheld boxes dashed, on the LiDAR (grey) and stereo (blue) points.}
  \label{fig:teaser}
\end{figure}
\begin{figure*}[t]
  \centering
  \begin{minipage}[t]{0.575\textwidth}\centering
    \IfFileExists{fig_rig_a.png}{\includegraphics[width=\linewidth]{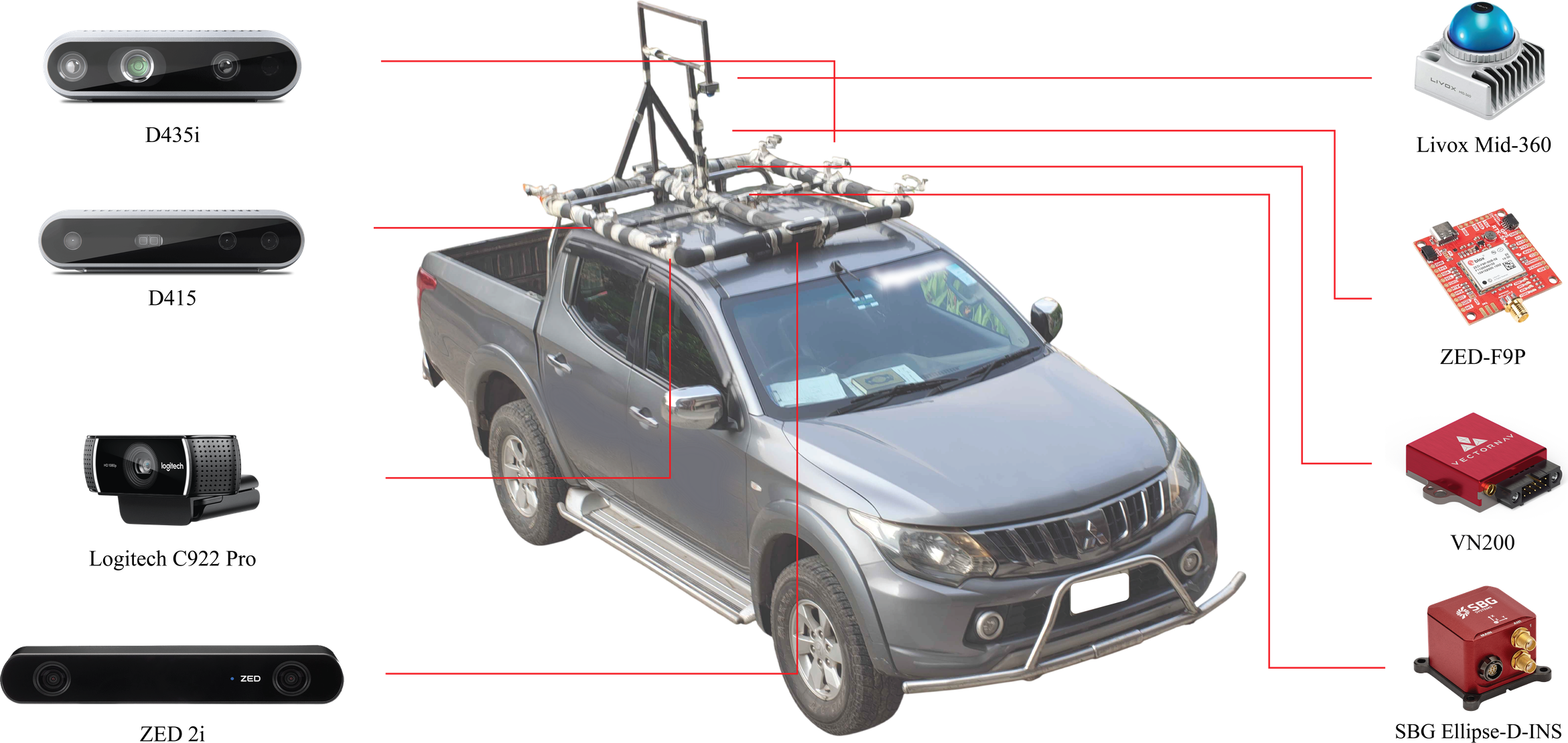}}{\phfig{4.7cm}{rig photo}}\\[1pt]\footnotesize (a)
  \end{minipage}\hfill
  \begin{minipage}[t]{0.415\textwidth}\centering
    \IfFileExists{fig_rig_b.png}{\includegraphics[width=\linewidth]{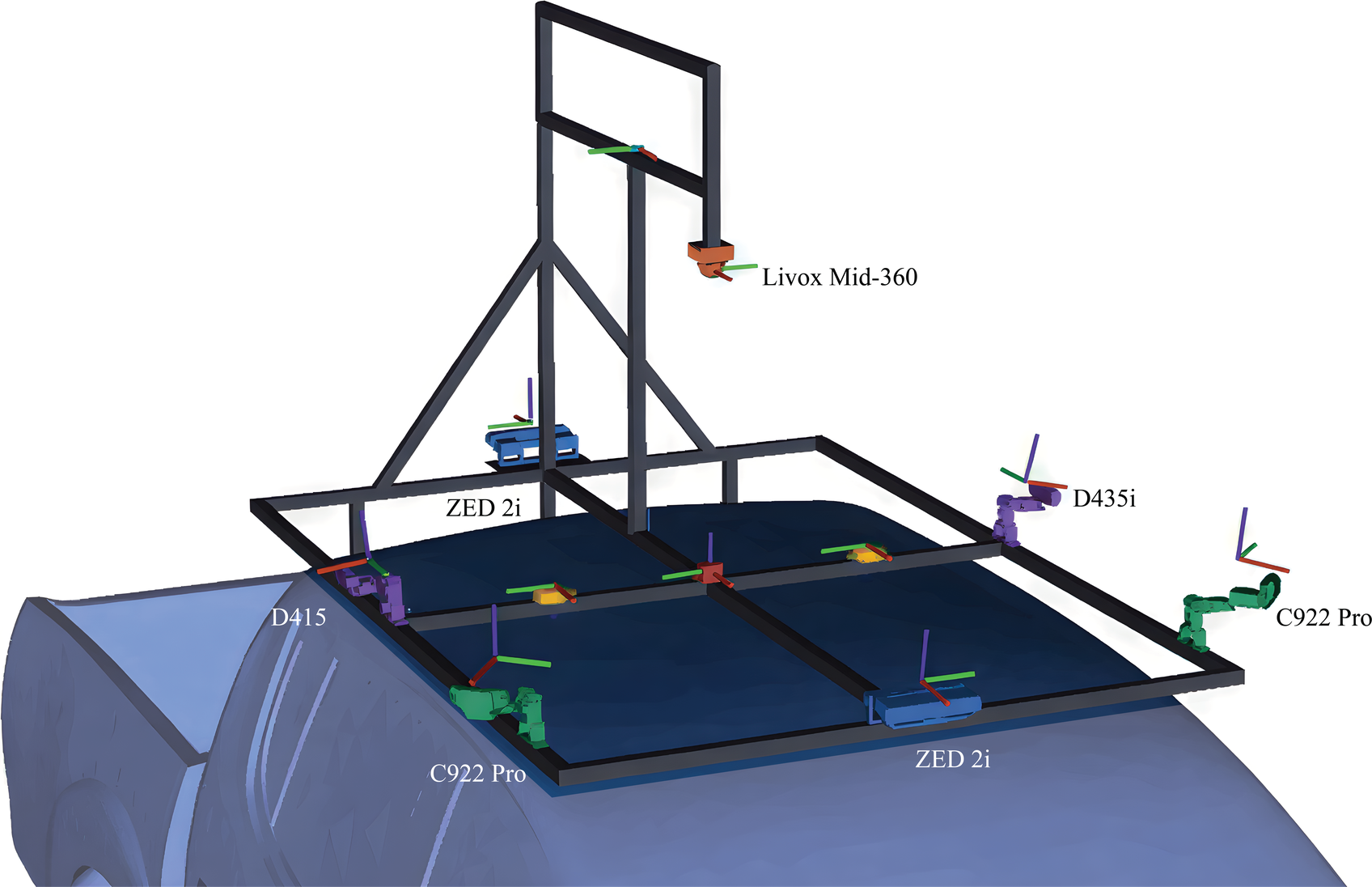}}{\phfig{4.7cm}{rig CAD}}\\[1pt]\footnotesize (b)
  \end{minipage}
  \caption{The \dsense{} sensor rig. (a) The sensors and their places on the roof deck of the vehicle, the Intel RealSense D435i and D415, the Logitech C922 Pro webcams, the two ZED~2i stereo cameras, the Livox Mid-360 LiDAR on the mast, the u-blox ZED-F9P GNSS receiver, the VectorNav VN-200 and the SBG Ellipse-D INS. (b) The CAD model of the rig with every sensor frame drawn as coordinate axes (x red, y green, z blue) and labelled, as released in the URDF.}
  \label{fig:rig}
\end{figure*}
This paper presents \dsense, a multimodal driving dataset and benchmark recorded in Dhaka, and makes four contributions.
\begin{enumerate}\setlength{\itemsep}{0pt}
  \item \textbf{Dense Traffic Multimodal Dataset:} \DatasetHours{} of synchronized LiDAR, stereo-depth, six-camera and GNSS/INS data in \DatasetScenes{} scenes of dense Dhaka traffic, with an \BenchNClasses-class taxonomy in which the CNG auto-rickshaw and cycle rickshaw are evaluated 3D categories with road labels. It is the first multimodal 3D driving dataset recorded in Bangladesh, combining LiDAR, stereo depth, cameras and GNSS/INS in one release (Sections~\ref{sec:sensors} and~\ref{sec:dataset}).
  \item \textbf{Density Stratified Benchmark:} An evaluation protocol that bins every keyframe by the measured density of road users within \DensityRadiusM\,\si{\metre} and by range, reports all metrics per stratum with a matched-class degradation, and uses route-disjoint splits (Section~\ref{sec:benchmark}).
  \item \textbf{Baselines:} We evaluate three-seed 3D detection baselines (PointPillars, CenterPoint, BEVFusion, and TransFusion-L) and point-wise segmentation on four fully annotated scenes. Six odometry methods are evaluated against a RTK reference trajectory, both with and without PPK correction. The rig’s own sensors are further evaluated as baseline measurements for depth estimation, occupancy mapping, and localization (Section~\ref{sec:experiments}).
  \item \textbf{Semiautomatic Human in the Loop Annotation Pipeline:} A pipeline that runs from a raw sensor export to gated 3D boxes, tracks and review tasks with no person in the loop, released with the capture stack and the benchmark toolkit (Section~\ref{sec:annotation}).
\end{enumerate}
\section{Related Work}
\label{sec:related}
\begin{figure*}[t]
  \centering
  \IfFileExists{fig_grid.pdf}{\includegraphics[width=\textwidth]{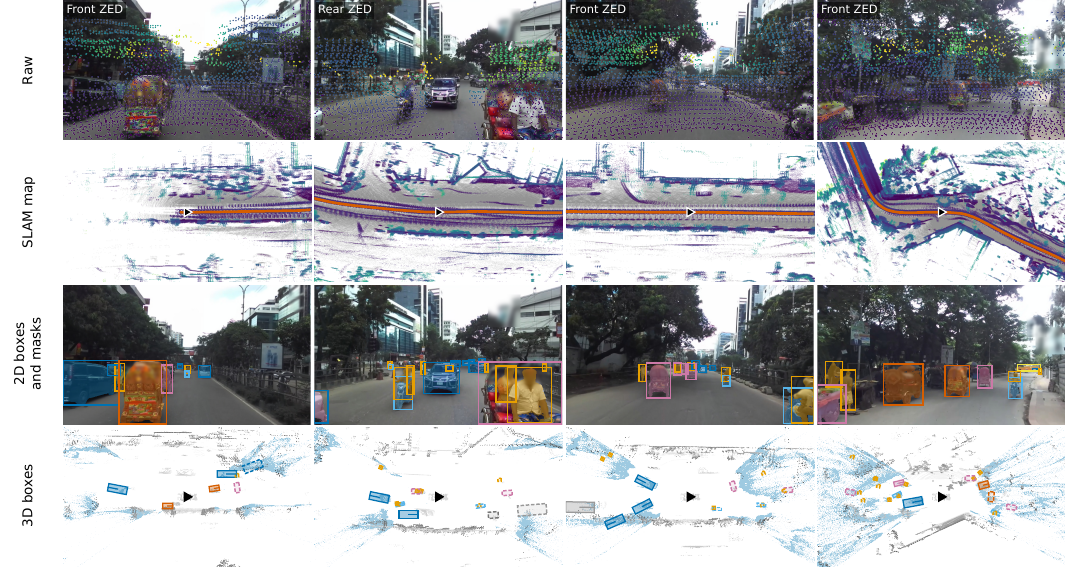}}{\phfig{4.6cm}{run make\_sample\_figs.py}}
  \caption{Data products for four keyframes of one \dsense{} scene, one keyframe per column. Top to bottom, the raw camera frame, front ZED unless labelled, with the LiDAR sweep projected and coloured by depth, the LiDAR map built by LiDAR odometry over the 80\,s around the keyframe with the trajectory in orange, the detector boxes and SAM~3.1 instance masks, and the 3D boxes on the LiDAR (grey) and stereo (blue) points, released boxes filled and boxes the quality gate withheld dashed. Both top-down rows are seen from above with forward to the right and the ego vehicle as the black triangle.}
  \label{fig:grid}
\end{figure*}
\noindent\textbf{Multimodal 3D Driving Datasets:}
The widely used multimodal benchmarks~\cite{geiger2012kitti,caesar2020nuscenes,sun2020waymo,wilson2021argoverse2,mao2021once} are not short of classes, since Argoverse~2 already has 26, but none includes a rickshaw or three-wheeler, and none pairs a low-cost non-repetitive LiDAR with a stereo and depth camera ring (Table~\ref{tab:datasets}).

\noindent\textbf{South Asian and Bangladeshi Datasets:}
The IDD family~\cite{varma2019idd,kumar2025driveindia} established unstructured-road perception in India with images, and IDD-3D~\cite{dokania2023idd3d} is the region's only multimodal 3D dataset.
Its 12k frames are daytime sequences from Hyderabad, and its ten trained categories name no rickshaw, so auto-rickshaws sit among untrained miscellaneous classes and cycle rickshaws are absent.
It has no night data, no odometry benchmark and no evaluation by traffic density.
Bangladesh has a growing image-only corpus~\cite{shihavuddin2020dhakaai,tabassum2020poribohon,zunair2024rsud20k,baig2024badodd} that names CNG, rickshaw and easy-bike classes but provides no LiDAR, 3D boxes, tracks or ego pose.

\noindent\textbf{Evaluation Protocols:}
We reuse the nuScenes detection kernels~\cite{caesar2020nuscenes} and the trajectory metrics of Sturm et al.~\cite{sturm2012benchmark} unchanged.
Stratified reporting has precedent per object~\cite{geiger2012kitti,sun2020waymo} and per synthetic corruption~\cite{kong2023robo3d}, whose mean-resilience-rate is the template for our Robustness Score, while scene-level density as an axis appears only in the 2D, single-class CrowdHuman~\cite{shao2018crowdhuman}.
Route-disjoint splits follow spatial block cross-validation~\cite{roberts2017cross}.

\noindent\textbf{Auto Annotation and Low Cost Rigs:}
Offboard labelling~\cite{qi2021offboard} refines with trained 3D models, and SAL~\cite{osep2024sal} lifts foundation-model output~\cite{carion2025sam3} into LiDAR zero-shot.
Our pipeline is the zero-shot branch of that recipe with a regionally fine-tuned detector for the classes no foundation model names.
Livox rigs with consumer cameras are precedented in MCD~\cite{nguyen2024mcd}, the Mid-360 has been characterized~\cite{mitka2025mid360}, calibration follows Koide et al.~\cite{koide2023calib}, and LiDAR-inertial odometry~\cite{xu2022fastlio2,zheng2025fastlivo2,koide2024glim} is both a baseline and a cross-check.

\section{Sensor Suite and Data Collection}
\label{sec:sensors}

\subsection{Platform and Sensors}
\begin{table}[t]
  \centering\footnotesize\setlength{\tabcolsep}{3pt}
  \caption{Sensor suite as built. Every camera records 720p at 30\,fps, rolling shutter, free running with host-clock timestamps.}
  \label{tab:sensors}
  \IfFileExists{table_sensors.tex}{\begin{tabular}{@{}llll@{}}
\toprule
Sensor & Role & Placement & H-FOV \\
\midrule
Livox Mid-360 & LiDAR, 5 Hz & Mast, inverted & 360$^\circ\times$59$^\circ$ \\
2$\times$ ZED 2i & Stereo & Front, back & 72$^\circ$ \\
2$\times$ C922 Pro & RGB & Front diagonals & 70.4$^\circ$ \\
1$\times$ RS D435i & RGB-D & Left & 69.4$^\circ$ \\
1$\times$ RS D415 & RGB-D & Right & 69.4$^\circ$ \\
u-blox ZED-F9P & GNSS timing & Deck & -- \\
SBG Ellipse-D & Primary INS, PPK & Deck & -- \\
VectorNav VN-200 & Backup INS and RTK & Deck & -- \\
\bottomrule
\end{tabular}
}{\tbd{run tools/dhakasense\_analyze.py}}
\end{table}
\dsense{} is recorded from the roof deck of a Mitsubishi L200 pickup (Fig.~\ref{fig:rig}, Table~\ref{tab:sensors}).
A single Livox Mid-360 LiDAR with a non-repetitive scan pattern is mounted \LidarInverted{} at \LidarHeightM\,\si{\metre} above the ground. Its coordinate system defines the rig reference frame and its timestamps the master time base.
Six cameras sit around the roof deck, \CamDropM\,\si{\metre} below the LiDAR, in the placements of Table~\ref{tab:sensors}.
The nominal horizontal fields of view in Table~\ref{tab:sensors} sum to \FovSumDeg\,\si{\degree}, a figure that includes overlapping views and is not unique angular coverage.
The release provides the LiDAR cloud and a single depth cloud fused from the two ZED~2i cameras in the ego frame.

For positioning and synchronization, a dedicated u-blox ZED-F9P receiver provides the timing reference for the LiDAR, while an \InsModel{} INS provides dual-frequency raw GNSS observations for post-processed kinematic (PPK) positioning and dual-antenna heading.
The dual-antenna heading avoids magnetometer or course-over-ground heading, which fails in stop-and-go traffic, and a VectorNav VN-200 is also recorded as an independent attitude reference.
The timing and PPK functions use separate receivers so that the raw-observation stream never disturbs the LiDAR timing pulse.

\subsection{Time Synchronization, Calibration and Ego Pose}
\label{sec:sync}
\begin{table*}[t]
  \centering\footnotesize
  \caption{Multimodal 3D and South Asian driving datasets compared with \dsense. Values from the cited papers. A dash means none and a question mark means not reported. $^\dagger$Pipeline labels, human verification in progress (Section~\ref{sec:tiers}).}
  \label{tab:datasets}
  \setlength{\tabcolsep}{2.7pt}
  \begin{adjustbox}{max width=\textwidth}
  \begin{tabular}{@{}llllcllccl@{}}
    \toprule
    Dataset & Region & LiDAR & Cams & Hours / frames & 3D boxes & Classes & Rickshaw type & Night & Licence \\
    \midrule
    KITTI~\cite{geiger2012kitti} & Karlsruhe & HDL-64E & 4 & 7.5k / 7.5k fr & 80k & 8 & -- & -- & CC BY-NC-SA \\
    Cityscapes~\cite{cordts2016cityscapes} & 50 cities, Germany & -- & 2 & 5k img & -- & 30 & -- & -- & Research only \\
    nuScenes~\cite{caesar2020nuscenes} & Boston, Singapore & HDL-32E & 6 & 5.5 h & 1.4M & 23 & -- & \checkmark & CC BY-NC-SA \\
    Waymo Open~\cite{sun2020waymo} & 6 US cities & 5 & 5 & 6.4 h & 12M & 4 & -- & \checkmark & NC \\
    Argoverse 2~\cite{wilson2021argoverse2} & 6 US cities & 2$\times$VLP-32C & 9 & 4.2 h & ? & 26 & -- & \checkmark & CC BY-NC-SA \\
    ONCE~\cite{mao2021once} & China & 40-beam & 7 & 144 h / 1M fr & 417k & 5 & -- & \checkmark & CC BY-NC-SA \\
    PandaSet~\cite{xiao2021pandaset} & San Francisco & Pandar64 + PandarGT & 6 & 0.2 h / 8.2k fr & ? & 28 & -- & \checkmark & CC0 \\
    A*3D~\cite{pham2020a3d} & Singapore & HDL-64E & 2 & 39k fr & 230k & 7 & -- & \checkmark & ? \\
    \midrule
    IDD-3D~\cite{dokania2023idd3d} & Hyderabad & OS1 & 6 & 15.5k fr & 223k & 17 & -- & -- & CC BY-SA \\
    RSUD20K~\cite{zunair2024rsud20k} & Bangladesh & -- & 1 & 20k img & -- & 13 & \checkmark & \checkmark & CC BY-NC-SA \\
    \midrule
    \textbf{\dsense} (ours) & Dhaka, Bangladesh & Mid-360 + stereo depth & \SessNCameras & \DatasetHoursDecimal{} h / \DatasetScenes{} scenes & \DatasetThreeDBoxesM$^\dagger$ & \BenchNClasses & \checkmark (3D) & -- & CC BY-NC-SA \\
    \bottomrule
  \end{tabular}
  \end{adjustbox}
\end{table*}
The F9P supplies a pulse per second and the NMEA \emph{GPRMC} sentence to the Mid-360, which stamps every point in GPS time, and \emph{chrony} disciplines the host clock from the same source to within \SyncOffsetUs\,\si{\micro\second} of LiDAR time.
Camera frames carry host arrival times, and the constant per-camera latency, measured after the first session by cross-correlating a moving target against the LiDAR to about $\pm$\LatencyResMs\,\si{\milli\second}, ships uncorrected for that session with the per-frame association residual recorded.
Intrinsics are checkerboard or factory values, camera-to-LiDAR extrinsics come from targetless direct visual-LiDAR calibration~\cite{koide2023calib} seeded from the measured mounts and re-run before every session, and every exported entry states whether it is factory or measured.
Ego poses come in three tiers, PPK fused with INS attitude, LiDAR-inertial odometry anchored to PPK epochs by Umeyama alignment~\cite{umeyama1991}, and odometry alone flagged as having no global reference, and FAST-LIO2~\cite{xu2022fastlio2} runs on every sequence as an independent check beside the PPK fix status.

\subsection{Collection Protocol}
\label{sec:collection}
\begin{figure}[t]
  \centering
  \IfFileExists{fig_map.pdf}{\includegraphics[width=0.75\columnwidth]{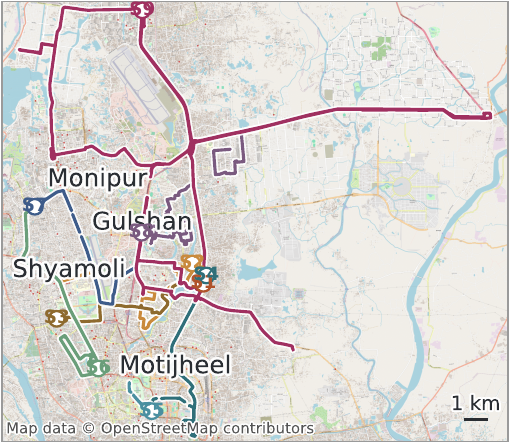}}{\phfig{6.0cm}{run make\_map\_fig.py}}
  \caption{Where \dsense{} was recorded. Trajectories over an OpenStreetMap base, colour coded with a numbered marker at each start and the most-visited areas named.}
  \label{fig:map}
\end{figure}

The recording sessions run from late afternoon into dusk and total \DatasetHours{} of driving in \DatasetScenes{} scenes of about five minutes each, exported at every LiDAR frame (\SessKeyframeHz\,\si{\hertz}) with intermediate camera frames as sweeps.
The sessions cover Motijheel and Rajarbag in the south-east, Rampura and Gulshan in the east and north-east, Tejgaon in the centre, and Shyamoli, Kolabagan, Mirpur and Monipur in the west and north-west (Fig.~\ref{fig:map}), with the recorder stopped between routes so that each session stays a separable route.

\section{The \dsense{} Dataset}
\label{sec:dataset}

\subsection{Taxonomy}
\label{sec:taxonomy}
\begin{table}[t]
  \centering\footnotesize\setlength{\tabcolsep}{3pt}
  \caption{\dsense{} class taxonomy with instances, boxes and share of instances per class on the four benchmark-release scenes of Section~\ref{sec:experiments}. $^\star$Headline long-tail classes.}
  \label{tab:taxonomy}
  \IfFileExists{table_classes.tex}{\begin{adjustbox}{max width=\columnwidth}\begin{tabular}{@{}llrrr@{}}
\toprule
Group & Class & Instances & Boxes & Share (\%) \\
\midrule
Motorised 4-wheel & Car & 7,357 & 29,776 & 25.0 \\
 & Truck & 846 & 3,151 & 2.9 \\
 & Bus & 602 & 2,064 & 2.0 \\
Indigenous 3-wheel & \textbf{CNG auto-rickshaw}$^{\star}$ & 1,220 & 6,807 & 4.2 \\
Human-powered & \textbf{Cycle rickshaw}$^{\star}$ & 1,129 & 5,861 & 3.8 \\
 & Bicycle & 1,058 & 2,347 & 3.6 \\
Motorised 2-wheel & Motorcycle & 2,610 & 9,048 & 8.9 \\
Vulnerable users & Pedestrian & 14,557 & 50,759 & 49.5 \\
\midrule
Total & 8 classes & 29,379 & 109,813 & 100.0 \\
\midrule
Road & LiDAR point layer & \multicolumn{3}{@{}r@{}}{\ifdefined\RoadShareLidarPct\RoadShareLidarPct\,\% of returns\else see text\fi} \\
\bottomrule
\end{tabular}
\end{adjustbox}}{\tbd{run tools/dhakasense\_analyze.py}}
\end{table}
Table~\ref{tab:taxonomy} lists the \NClasses{} object classes of the full \dsense{} taxonomy.
Classes are grouped by acceleration and steering rather than by physical size, since in mixed traffic the way a vehicle is driven, not its size, governs its speed, turning radius and the gaps it accepts.
The \NClasses{} classes fall into five groups, motorised four-wheelers, indigenous three-wheelers, human-powered vehicles, motorised two-wheelers and vulnerable road users.
Six classes map one-to-one onto nuScenes detection classes, so models and metrics built for nuScenes transfer without remapping.
Boxes are annotated when at least some LiDAR point falls inside them and the object is visible in some camera, matching the nuScenes rule, and each carries a visibility level, an attribute and its LiDAR point count.

\subsection{Coverage}
The \DatasetScenes{} scenes come from one city (Section~\ref{sec:collection}), on routes chosen to span the density spectrum rather than distance.
The campaign has \DatasetSessions{} sessions and \DatasetKm\,\si{\kilo\metre} of driving through \DatasetAreas{} named areas, each with a post-processed GNSS solution.

\subsection{Splits and Label Tiers}
\label{sec:tiers}
\textbf{Label Tiers:} Every keyframe belongs to one of three label tiers, recorded per box so that a user can select the label quality an experiment requires.
The \emph{verified} tier holds human-corrected 3D boxes, track identities and instance masks.
The \emph{pipeline} tier holds boxes produced by the annotation pipeline of Section~\ref{sec:annotation}, each with the confidence tier the quality gate assigned to it.
The \emph{unverified} tier holds pipeline output that no reviewer has seen.
Boxes that the gate withheld are shipped in a side table with the reason for their exclusion, so recall against the raw detector output can be reconstructed.

\textbf{State of the Release:} The label files cover the \ProcessedHours{} of sessions processed so far, all in the pipeline tier. Four of these scenes form the benchmark release (Section~\ref{sec:experiments}), with \RelNAnnotationsK{} boxes in Table~\ref{tab:taxonomy}. The gate admitted \DatasetAdmittedM{} boxes (\DatasetAdmittedPct\,\% of all gated boxes) and withheld \DatasetWithheldM{} (\DatasetWithheldPct\,\%), of which \DatasetFlaggedM{} (\DatasetFlaggedPct\,\%) are flagged for review and \DatasetRejectedM{} (\DatasetRejectedPct\,\%) rejected.

\subsection{Release Format, Privacy and Licence}
\label{sec:release}
\dsense{} releases the raw sensor data of all \DatasetScenes{} scenes, \DatasetHours{} of driving with \RawLidarPointsB{} billion LiDAR returns, whereas the labels cover the \ProcessedHours{} of sessions processed so far, \DatasetThreeDBoxesM{} 3D boxes and \DatasetRoadPointsM{} million road-labelled points, in the nuScenes table format, with every LiDAR frame as a keyframe and camera sweeps between keyframes, so that existing nuScenes tooling can be reused.
The road layer ships in both of its forms, the camera road masks of every frame and the per-point road labels of every sweep, next to the per-point object labels that the painted masks give the LiDAR cloud.
Each sequence comes with a sensor-benchmark quality record that gives the time-sync report, the calibration residuals, the PPK fix ratio and, for every calibration value, whether it was measured or nominal.
Faces and vehicle licence plates are blurred in all released images after annotation.
The data are released under CC~BY-NC-SA~4.0 and the capture stack, annotation pipeline and benchmark toolkit under GPL-3.0.
Labels descended from the fine-tuned detector arm inherit the CC~BY-NC~4.0 terms of its training set and are marked per box.
The capture stack, annotation pipeline and benchmark toolkit are available at \url{https://anonymous.4open.science/r/so_sleepy-C186}.

\section{Semiautomatic Human in the Loop Annotation Pipeline}
\label{sec:annotation}
\begin{figure*}[!t]
  \centering
  \IfFileExists{fig_pipeline_v4.pdf}{\includegraphics[width=\textwidth]{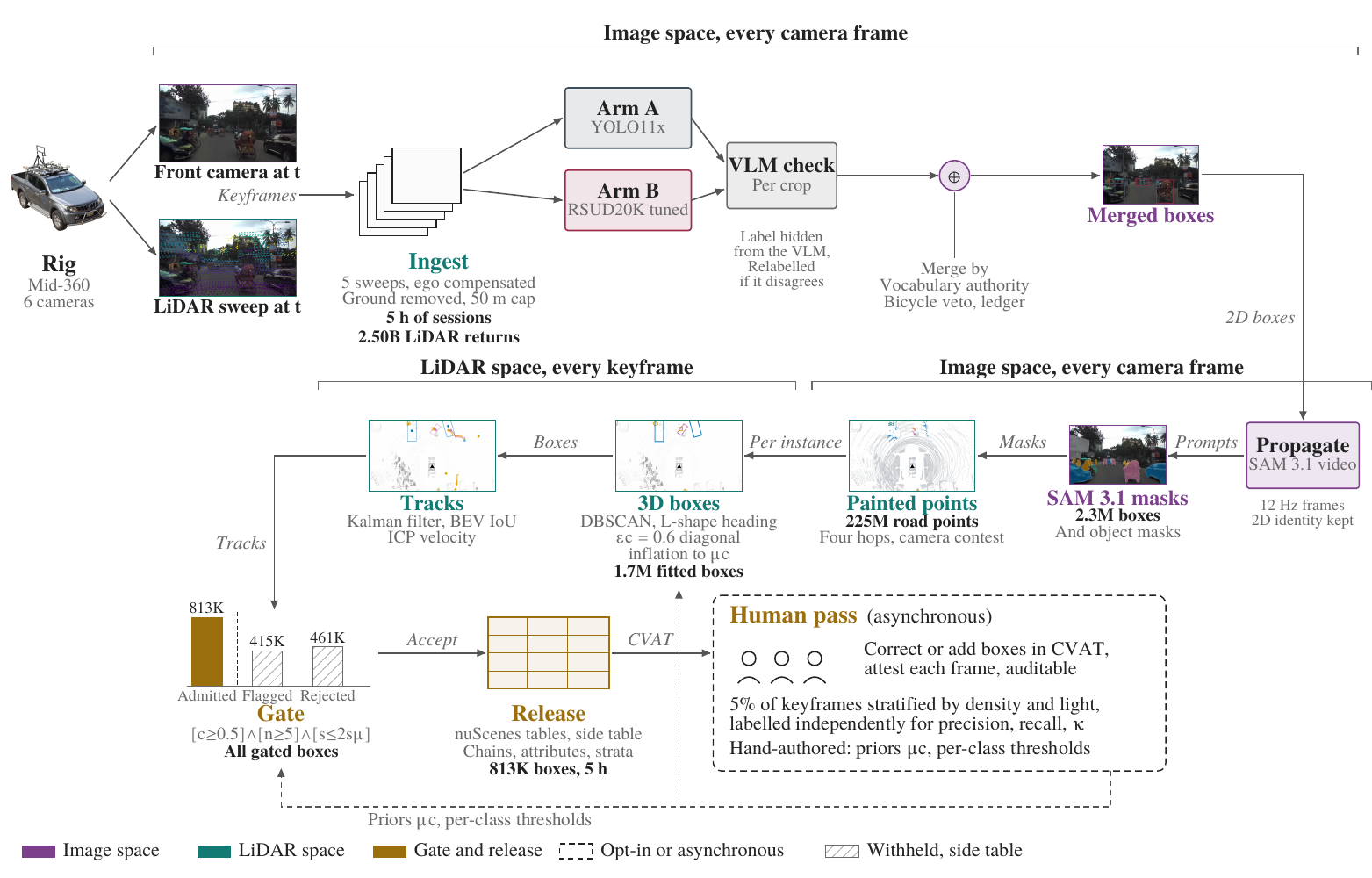}}{\figtbd{run make\_pipeline\_v4.py}}
  \caption{The annotation pipeline, illustrated on one released keyframe with counts for the \ProcessedHours{} of sessions processed so far. Following the arrows across the three rows, the rig and ingestion, the two detector arms, the blind VLM check and the merge, SAM~3.1 propagation and masks, painting into the LiDAR, box fitting and tracking, the gate, the release and the asynchronous human pass, which feeds priors and thresholds back. The legend gives the colour key, and bold counts are totals over the processed sessions.}
  \label{fig:pipeline}
\end{figure*}
Annotating 3D boxes by hand at nuScenes scale is beyond a university group's budget, so \dsense{} labels are produced by a pipeline that runs from a raw sensor export to gated 3D boxes, tracks and review tasks without a person in the loop (Fig.~\ref{fig:pipeline}), and people verify and correct afterwards.
Every stage exits clean, degraded with its causes named, or refused with nothing written, and writes a manifest binding its output to a fingerprint of the substrate, configuration and model revisions, so a crash can never pass as a quality flag, and every box carries its provenance and gate vector.

\subsection{Stages}
\subsubsection{Substrate}
A probe admits a scene through six named predicates on version, channel completeness, file parsing, sweep coverage and token-graph closure, and excludes a failing scene with the predicate named.
Ingestion applies calibration once, subsamples the two stereo depth clouds the exporter fused into the sweep to every eighth point, drops the rear stereo cloud because its extrinsic tilts its road plane about \SI{4}{\degree} against the LiDAR's, removes ground with a sector-wise RANSAC~\cite{fischler1981ransac} plane fit and caps range at \SI{50}{\metre}.

\subsubsection{Two Arm 2D Proposals, VLM Verification, Merge and 2D Tracking}
No COCO detector can name a rickshaw, and open-vocabulary detectors cross-fire between phrases sharing a head noun, so two arms run on every frame.
Arm~A is YOLO11x~\cite{khanam2024yolo11} with its COCO vocabulary, and arm~B is the same architecture fine-tuned on RSUD20K~\cite{zunair2024rsud20k}, from which only the rickshaw and CNG auto-rickshaw outputs are kept, since COCO lacks these locally prevalent classes.
Before merging, a blind vision-language model, Nemotron 3 Nano Omni~\cite{nvidia2026nemotron3nanoomni}, checks the bounding-box crops from both arms to correct class mislabels and filter false positives.
The arms are merged by a fixed authority table rather than by score, so an arm-B rickshaw suppresses an overlapping arm-A car, truck, bus, motorcycle or bicycle, a confident arm-A bicycle vetoes the arm-B box instead, and a rider box mostly inside a rickshaw is absorbed.
Nothing is deleted, and a suppressed box moves to a ledger on the row that beat it.
The merged boxes are then passed to the SAM~3.1 video tracker, the v3.1 release of SAM~3~\cite{carion2025sam3}, to propagate keyframe boxes through the \SI{12}{\hertz} frames.

\subsubsection{Masks and Painting}
Each box prompts SAM~3.1 for an instance mask, duplicates across overlapping cameras are suppressed by angular overlap in the ego frame, and every LiDAR point is projected into every camera through the ego and global frames at the two timestamps~\cite{vora2020pointpainting}, so that the free-running cameras are compensated for ego motion, and awarded to the camera whose optical axis it lies closest to.
A road stage labels the sweep as well: a LiDAR point that projects into a camera road mask is kept as road when it lies on the sector-wise ground plane and rejected off-plane, which marks \RoadShareLidarPct\,\% of the returns it sees and \RoadShareSeenPct\,\% of those a firing camera covers, about \RoadPtsPerKf{} points per keyframe. Because an unmarked carriageway offers no lane geometry to reason from, this layer is what lets a released box be tested for whether its obstacle stands on the road or beside it.

\subsubsection{Boxes and Tracks}
Painted points are clustered in the bird's-eye view by DBSCAN~\cite{ester1996dbscan} with a class-conditional radius from the class prior, and the largest cluster is fitted with an L-shape by a brute-force heading search, flagging near-square or sparse boxes as yaw ambiguous.
A constant-velocity Kalman filter with Hungarian association over bird's-eye IoU, refined by ICP between adjacent keyframes, links boxes into tracks~\cite{weng2020ab3dmot}.
A box with fewer than 15 returns is inflated toward the class-mean dimensions by a clamped one-sided blend that anchors the sensor-facing face and the top, never shrinks and never exceeds the class mean, and the measured box is retained beside the shipped one.

\subsubsection{Quality Gate and Release}
The gate evaluates three predicates on the pre-inflation box,
\begin{equation}
  g_1 = [\,c \ge 0.5\,],\;
  g_2 = [\,n \ge 5\,],\;
  g_3 = [\,s \le 2 s_{\mu} \wedge \ell \le 2 \ell_{\mu}\,],
  \label{eq:gate}
\end{equation}
where $c$ is detector confidence, $n$ the single-sweep return count, and $s,\ell$ and $s_\mu,\ell_\mu$ the footprint sides of the box and of the class prior.
A box is rejected if $g_2$ or $g_3$ fails, flagged if $g_1$ fails or no prior exists, and otherwise admitted at the auto-accept tier, and a tier never upgrades the record's source, which stays \emph{pipeline}.
The release writer stitches identities across gaps of up to \StitchMaxGap{} keyframes, interpolates the gaps, assigns motion attributes at \MovingSpeedMps\,\si{\metre\per\second}, refuses any category it cannot map, and publishes review tasks to CVAT~\cite{cvat2020} additively under a run tag.

\subsection{Reach, Computational Cost and the Human Boundary}
\label{sec:humanboundary}
Processing \ProcessedHours{} of sessions required \ProcessedGpuHours{} GPU-hours on an NVIDIA GeForce RTX 4090 with 24\,GB of VRAM, and each scene contains approximately 40K annotations, for a total of \DatasetBoxesM{} annotated 2D bounding boxes and \DatasetThreeDBoxesM{} 3D boxes.
Humans enter at four points, namely review of the published CVAT tasks with per-frame attestation, authoring of the class dimension priors, per-class threshold tuning, and a stratified \DoubleFractionPct\,\% of keyframes that is fully annotated by humans and used to validate the released labels through precision, recall and inter-annotator agreement (3D IoU and Cohen's $\kappa$~\cite{cohen1960kappa}).
At submission the human pass had been published but had not completed, so every released box remains pipeline output, and annotator hours, precision, recall and $\kappa$ are not yet available.

\section{Density Stratified Benchmark}
\label{sec:benchmark}
\begin{table*}[t]
\centering
\begin{minipage}[t]{0.485\textwidth}\centering\footnotesize
  \captionof{table}{3D detection and point segmentation, mean over four scenes and three seeds.}\label{tab:daytwo}
  {\setlength{\tabcolsep}{3pt}\IfFileExists{table_daytwo.tex}{\begin{adjustbox}{max width=\linewidth}\begin{tabular}{@{}llrrrrrr@{}}
\toprule
& & \multicolumn{5}{c}{3D detection} & Segm. \\
\cmidrule(lr){3-7}\cmidrule(lr){8-8}
Method & Mod. & mAP & Rickshaw AP & mATE (m) & mAOE (rad) & NDS & mIoU \\
\midrule
CenterPoint~\cite{yin2021centerpoint} & S & \textbf{38.7} & \textbf{54.2} & \textbf{0.43} & 1.39 & \textbf{36.7} & \textbf{17.7} \\
BEVFusion~\cite{liu2023bevfusion} & S+C & 31.3 & 48.5 & 0.53 & 1.30 & 29.5 & 9.6 \\
PointPillars~\cite{lang2019pointpillars} & S & 24.1 & 40.5 & 0.58 & \textbf{1.26} & 26.8 & 11.5 \\
TransFusion-L~\cite{bai2022transfusion} & S & 21.7 & 23.7 & 0.55 & 1.40 & 24.7 & 6.3 \\
\bottomrule
\end{tabular}
\end{adjustbox}}{\tbd{run the harness}}}
\end{minipage}\hfill
\begin{minipage}[t]{0.485\textwidth}\centering\footnotesize
  \captionof{table}{Per-class AP in percent, mean over three seeds and four scenes.}\label{tab:perclass}
  \IfFileExists{table_perclass_main.tex}{\begin{adjustbox}{max width=\linewidth}\begin{tabular}{@{}lrrr@{}}
\toprule
Class & CenterPoint & BEVFusion & PointPillars \\
\midrule
Cycle rickshaw & \textbf{72.6} & 61.4 & 61.1 \\
Motorcycle & \textbf{70.5} & 50.6 & 0.4 \\
Pedestrian & \textbf{63.4} & 58.5 & 51.1 \\
Car & \textbf{49.1} & 29.7 & 45.4 \\
CNG auto-rickshaw & \textbf{35.7} & 35.5 & 19.9 \\
Truck & \textbf{17.3} & 13.0 & 13.6 \\
\bottomrule
\end{tabular}
\end{adjustbox}}{\tbd{run the harness}}
\end{minipage}

\vspace{0pt}
\begin{minipage}[t]{0.485\textwidth}\centering\footnotesize
  \captionof{table}{NDS per density stratum, percent. $\Delta^{\uparrow}_\rho$ is NDS(Extreme) minus NDS(Low), so positive means better when crowded. $\dagger$ thin stratum.}\label{tab:strata}
  \IfFileExists{table_strata.tex}{\begin{adjustbox}{max width=\linewidth}\begin{tabular}{@{}lrrrrrr@{}}
\toprule
& \multicolumn{4}{c}{NDS by density bin} & \multicolumn{2}{c}{$\Delta^{\uparrow}_\rho$} \\
\cmidrule(lr){2-5}\cmidrule(lr){6-7}
Method & Low & Medium & High & Extreme & Raw & Matched \\
\midrule
CenterPoint & 30.2$^{\dagger}$ & 36.2 & 35.6 & 36.5 & +6.3 & +4.1 \\
BEVFusion & 23.4$^{\dagger}$ & 29.2 & 29.3 & 29.8 & +6.4 & +6.3 \\
PointPillars & 21.9$^{\dagger}$ & 25.8 & 26.1 & 26.0 & +4.1 & +1.8 \\
TransFusion-L & 19.7$^{\dagger}$ & 23.4 & 24.5 & 26.0 & +6.3 & +5.6 \\
\bottomrule
\end{tabular}
\end{adjustbox}}{\tbd{run the harness}}
\end{minipage}\hfill
\begin{minipage}[t]{0.485\textwidth}\centering\footnotesize
  \captionof{table}{ATE in m per scene against the PPK reference, without and with the PPK solution fused as a position input. L LiDAR, I IMU, S stereo, C camera. A dash marks a scene without a valid result.}\label{tab:odom}
  {\setlength{\tabcolsep}{2.5pt}\IfFileExists{table_odom.tex}{\begin{adjustbox}{max width=\linewidth}\begin{tabular}{@{}llrrrrrrrr@{}}
\toprule
& & \multicolumn{4}{c}{ATE (m)} & \multicolumn{4}{c}{ATE with PPK (m)} \\
\cmidrule(lr){3-6}\cmidrule(lr){7-10}
Method & Sensors & F & G & H & I & F & G & H & I \\
\midrule
KISS-ICP~\cite{vizzo2023kissicp} & L & 4.9 & 6.0 & 2.9 & 5.7 & \textbf{1.4} & 1.5 & 1.9 & \textbf{2.6} \\
GLIM~\cite{koide2024glim} & L+I & \textbf{3.6} & \textbf{4.0} & \textbf{2.3} & \textbf{3.3} & \textbf{1.4} & 1.5 & \textbf{1.8} & 3.2 \\
LIO-SAM~\cite{shan2020liosam} & L+I & -- & 5.1 & 2.6 & -- & -- & 1.4 & 2.1 & -- \\
FAST-LIVO2~\cite{zheng2025fastlivo2} & L+I+C & -- & 4.1 & 2.8 & -- & -- & 1.6 & 1.8 & -- \\
\midrule
ZED VIO (front) & S+I & 18.2 & 15.5 & 7.8 & 45.2 & 6.9 & \textbf{1.4} & 1.9 & 2.9 \\
ZED VIO (rear) & S+I & 8.7 & 19.9 & 10.8 & 40.8 & 6.8 & 1.7 & 2.0 & 3.5 \\
ORB-SLAM3~\cite{campos2021orbslam3} (front) & S+I & -- & 8.4 & 8.6 & 4.5 & -- & 4.1 & 2.8 & 3.5 \\
\bottomrule
\end{tabular}
\end{adjustbox}}{\tbd{run the harness}}}
\end{minipage}
\end{table*}
Our benchmark wraps the official nuScenes~\cite{caesar2020nuscenes} and KITTI~\cite{geiger2012kitti} metric kernels unchanged, adds stratified reporting, and stamps every result file with the benchmark definition's hash.

\textbf{Tasks and Metrics:} 3D detection follows the nuScenes protocol, with centre-distance mAP over the eight classes, the true positive errors and NDS, and AP over the two rickshaw classes as a separate headline. Point segmentation is scored as mIoU over the classes a method is eligible for, on two label layers, the eight object classes on the stereo cloud and road on the LiDAR sweep. Stereo depth is scored against the LiDAR projection with AbsRel, RMSE and $\delta<1.25^{k}$~\cite{eigen2014depth}, and occupancy against a $\pm4$ keyframe accumulation with voxel mIoU and RayIoU~\cite{liu2024sparseocc}. Odometry follows KITTI, with $t_{\mathrm{rel}}$ and $r_{\mathrm{rel}}$ over 100 to \SI{800}{\metre} sub-sequences and ATE after SE(3) alignment~\cite{sturm2012benchmark}, against the post-processed kinematic (PPK) GNSS trajectory at \SI{1}{\hertz}. An estimator that consumes the PPK solution is a localization system and is scored on its along-track and cross-track error separately.

\label{sec:density}
\textbf{Strata:} For each keyframe $t$ the road-user density is
\begin{equation}
  \dens_t = \frac{N_t(R)}{\pi R^{2}},\qquad
  N_t(R) = \bigl|\{\, i : \lVert \mathbf{p}_{i,t} - \mathbf{p}^{\mathrm{ego}}_{t} \rVert_2 \le R \,\}\bigr|,
  \label{eq:density}
\end{equation}
with $\mathbf{p}_{i,t}$ the bird's-eye-view centre of annotated agent $i$ and $R=\DensityRadiusM\,\si{\metre}$, binned into Low, Medium, High and Extreme at the \DensityQuantiles{}\,\% quantiles of the dataset's own distribution. Range is binned by distance band. Every metric is reported per bin with its change from Low to Extreme, raw and matched over the classes present in both bins.

\label{sec:splits}
\textbf{Splits:} Benchmark splits are route-disjoint, formed by clustering keyframe trajectories with a \RouteSepM\,\si{\metre} proximity union-find and assigning whole clusters to train, val and test~\cite{roberts2017cross}. The verified tier forms validation and test. The benchmark is evaluated on 4 scenes, so every baseline below uses a provisional block-random split on which no leaderboard is opened, and test labels are withheld.

\section{Experiments}
\label{sec:experiments}
\label{sec:setup}
We benchmark on four annotated scenes, which ship the Mid-360 sweep and the stereo cloud as separate channels and carry boxes of class-prior length with fitted width and height. Detectors are trained with OpenPCDet~\cite{openpcdet2020} from nuScenes initializations on a pooled block-random split of the four scenes, three seeds each, and evaluated on each scene's validation slice.

\label{sec:det}
\textbf{3D Detection and Segmentation:} CenterPoint leads with \DayTwoCenterPointMeanmAP\,\% mAP and \DayTwoCenterPointMeanRickshawAP\,\% AP on the rickshaw classes, its seed spread never exceeding \DayTwoCenterPointMaxSeedSd{} points (Table~\ref{tab:daytwo}), BEVFusion follows at half the epochs, PointPillars trails, and TransFusion-L had not converged at the common budget. The rickshaw classes are detected better than the average class. A single anchor set covers \DayTwoAnchorCoveragePct\,\% of the class-prior boxes at the strictest matching threshold. Every class has constant measurement and mASE reports the error from that constant. Per class (Table~\ref{tab:perclass}), the cycle rickshaw and the motorcycle are the best detected classes for CenterPoint and the CNG auto-rickshaw is within reach of the car. In the box-derived point segmentation the cameras help only on pedestrians, where BEVFusion is best.

\label{sec:stratified}

\textbf{Strata:} Accuracy does not fall with crowding (Table~\ref{tab:strata}). All four detectors score higher at Extreme than at Low density, by \DensDeltaCenterPoint{} NDS points for CenterPoint, in the raw and the matched form. Density on this route is a proxy for range, since a sparse keyframe is an open stretch whose few objects are distant, and every model loses several points from the 0 to \SI{20}{\metre} band to the 20 to \SI{40}{\metre} band. The Low bin holds \StrataGtLow{} objects against \StrataGtExtreme{} in Extreme and is the noisiest estimate, and the band beyond \SI{40}{\metre} holds no ground truth.

\label{sec:odom}

\textbf{Odometry:} The reference has \PpkDayTwoFixedPct\,\% RTK-fixed and \PpkDayTwoFloatPct\,\% RTK-float epochs with a horizontal sigma of \PpkSigmaFixedMm{} and \PpkSigmaFloatMm\,\si{\milli\metre}, and its heading agrees with its own course to \PpkHeadingAgreeDeg\,\si{\degree}. GLIM and KISS-ICP hold track on every scene, GLIM between \OdomGLIMMinATE{} and \OdomGLIMMaxATE\,\si{\metre} ATE and KISS-ICP between \OdomKISSICPMinATE{} and \OdomKISSICPMaxATE\,\si{\metre} (Table~\ref{tab:odom}), and the tightly coupled LiDAR-inertial systems match or beat them on the scenes they survive, a recording gap and a road shock separating the two groups. Fusing the PPK solution brings every estimator that held track to between \PpkCorrMinATE{} and \PpkCorrMaxATE\,\si{\metre}, so the trajectory is accurate to about two metres over the drive.

\label{sec:rigbaselines}
\textbf{Depth, Occupancy and Localization:} No trained baseline exists for these tasks yet, so the rig's own sensors set the first mark. Stereo depth against the LiDAR projection reaches \DepthAbsRelMean{} AbsRel after an 8-pixel fill of the reprojected cloud. A single sweep scores \OccVoxelMean\,\% voxel mIoU but \OccRayMean\,\% RayIoU, because it fills \OccSweepFillPct\,\% of the grid against \OccAccumFillPct\,\% for the accumulation while its first-hit surfaces are right three times in four, so the gap for an occupancy method is sampling rather than accuracy. The PPK-fused LiDAR estimators are close to sub-metre across the lane and \LocGLIMRatio{} to \LocKISSICPRatio{} times worse along it, the drivable failure direction.

\section{Limitations}
\label{sec:limitations}
\dsense{} is a single-city, single-vehicle dataset whose benchmarked scenes cover one afternoon route, and it cannot claim to represent South Asia or even Dhaka's daytime traffic.
The benchmark covers a very small subset of the dataset because of the computational constraints we had.
We chose consumer-grade cameras so that any group in the region can replicate the rig, and the price is paid in sensor quality, since all cameras are rolling shutter and free running without hardware triggering, so residual timing error will limit camera-based 3D methods.
Every released label is pipeline output pending human verification, so the baseline numbers measure agreement with the pipeline, and no route-disjoint split exists yet, so every split is provisional.

\section{Conclusion}
We presented \dsense, a multimodal driving dataset and benchmark for dense, heterogeneous traffic, recorded with a low-cost rig.
Its \NClasses-class taxonomy names the indigenous three-wheelers that dominate South Asian roads, and its density-stratified protocol turns the claim that \emph{our data is harder} into a measured curve.
On the four benchmarked scenes the best detector reaches \DayTwoCenterPointMeanmAP\,\% mAP, well below the \SI{60}{\percent} the same models reach on nuScenes, and accuracy improves rather than degrades as the scene grows more crowded, which points to range rather than density as the dominant difficulty.
LiDAR odometry stays within \OdomGLIMMinATE{} to \OdomGLIMMaxATE\,\si{\metre} of the PPK reference over kilometre-long scenes and within \PpkCorrMinATE\,\si{\metre} once the PPK solution is fused. The released road masks and the per-point road and object labels extend the benchmark beyond boxes, towards drivable-area estimation and road-condition heuristics for streets that carry no lane markings to read.

\section*{Acknowledgment}
We thank BRAC University for supporting this work and the Dhaka Metropolitan Police (DMP) for their cooperation during data collection.

\bibliographystyle{IEEEtran}
\bibliography{IEEEabrv,refs}

\clearpage
\appendices
\section*{Appendix}
These appendices carry what the page budget of the main paper excluded, in the order a reader is likely to want it. Section~\ref{supp:comparison} gives the complete dataset comparison, including the image-only South Asian sets. Section~\ref{supp:sensors} gives the full sensor table and the time synchronization, calibration and ego-pose procedures. Section~\ref{supp:collection} describes the collection protocol, the planned districts and every recorded session. Section~\ref{supp:taxonomy} gives the full taxonomy, the label tiers and the class statistics of the earlier fused-cloud scenes. Section~\ref{supp:annotation} is the stage-by-stage pipeline reference with every threshold as implemented. Section~\ref{supp:release} documents the release format and its known deviations. Section~\ref{supp:benchmark} gives the benchmark protocol in full and Section~\ref{supp:experiments} the complete baseline tables. Section~\ref{supp:roadmap} lists what remains to be done. Numbers are produced by the same analysis harness as the main paper.
\section{Dataset Comparison in Full}
\label{supp:comparison}
\begin{table*}[t]
  \centering\footnotesize
  \caption{Multimodal 3D driving datasets and South Asian driving datasets compared with \dsense. Values from the cited papers. A dash means none and a question mark means not reported. $^\ddagger$pipeline labels awaiting human verification.}
  \label{stab:datasets}
  \setlength{\tabcolsep}{3pt}\scriptsize
  \begin{adjustbox}{max width=\textwidth}
  \begin{tabular}{@{}llllcllccl@{}}
    \toprule
    Dataset & Region & LiDAR & Cams & Hours / frames & 3D boxes & Classes & Rickshaw type & Night & Licence \\
    \midrule
    \multicolumn{10}{@{}l}{\textit{Multimodal 3D}} \\
    KITTI~\cite{geiger2012kitti} & Karlsruhe & HDL-64E & 4 & 7.5k fr & 80k & 8 & -- & -- & CC BY-NC-SA \\
    nuScenes~\cite{caesar2020nuscenes} & Boston, Singapore & HDL-32E & 6 & 5.5 h & 1.4M & 23 & -- & \checkmark & CC BY-NC-SA \\
    Waymo Open~\cite{sun2020waymo} & 6 US cities & 5 & 5 & 6.4 h & 12M & 4 & -- & \checkmark & NC \\
    Argoverse 2~\cite{wilson2021argoverse2} & 6 US cities & 2$\times$VLP-32C & 9 & 4.2 h & ? & 26 & -- & \checkmark & CC BY-NC-SA \\
    ONCE~\cite{mao2021once} & China & 40-beam & 7 & 144 h (16k annot.) & 417k & 5 & -- & \checkmark & CC BY-NC-SA \\
    IDD-3D~\cite{dokania2023idd3d} & Hyderabad & OS1 & 6 & 15.5k fr & 223k & 17 & -- & ? & CC BY-SA \\
    PHE3D~\cite{loc2025phe3d} & Vietnam & yes & -- & ? & ? & 5 & -- & ? & ? \\
    \midrule
    \multicolumn{10}{@{}l}{\textit{Image only, South Asia}} \\
    IDD~\cite{varma2019idd} & India & -- & 1 & 10k img & -- & 34 & \checkmark & ? & research \\
    DriveIndia~\cite{kumar2025driveindia} & India & -- & ? & 67k img & -- & 24 & \checkmark & -- & CC BY \\
    DhakaAI~\cite{shihavuddin2020dhakaai} & Dhaka & -- & ? & 3k img & -- & 21 & \checkmark & ? & CC0 \\
    Poribohon-BD~\cite{tabassum2020poribohon} & Bangladesh & -- & ? & 9k img & -- & 15 & \checkmark & ? & CC BY \\
    RSUD20K~\cite{zunair2024rsud20k} & Bangladesh & -- & 1 & 20k img & -- & 13 & \checkmark & \checkmark & CC BY-NC-SA \\
    BadODD~\cite{baig2024badodd} & Bangladesh & -- & 1 & 9.8k img & -- & 13 & \checkmark & \checkmark & CC BY \\
    \midrule
    \textbf{\dsense} (ours, earlier fused-cloud scenes) & Dhaka, Bangladesh & Mid-360 + stereo depth & \SessNCameras & \SessMinutes{} min / \RelNSamples{} kf & \RelNAnnotations & \BenchNClasses & \checkmark (3D) & dusk$^\ddagger$ & CC BY-NC-SA \\
    \bottomrule
  \end{tabular}
  \end{adjustbox}
\end{table*}
Table~\ref{stab:datasets} restores the rows that the main paper's comparison table dropped for space.
The Western multimodal sets differ from \dsense{} in three ways that matter here.
None includes a rickshaw or three-wheeler class, even where the taxonomy is deep, as in Argoverse~2 with 26 classes.
All use spinning LiDARs, whereas \dsense{} uses a non-repetitive solid-state unit fused with stereo depth.
Their densest scenes are far below Dhaka's mixed traffic, which is the property the benchmark's density axis is built to expose.
The South Asian image-only sets name the local classes and are the source of the fine-tuned detector arm, but none provides LiDAR, 3D boxes, tracks or ego pose.

\section{Sensor Suite Details}
\label{supp:sensors}

\subsection{Time Synchronization}
\label{supp:sync}
The F9P supplies a pulse per second and the NMEA \emph{GPRMC} time sentence to the Mid-360, which stamps every point in GPS time, and the host clock is disciplined by \emph{chrony} from the same source.
On the bench this path holds the host clock \SyncOffsetUs\,\si{\micro\second} from LiDAR GPS time with a \SyncStdUs\,\si{\micro\second} standard deviation, and every clock transition during a recording is logged so that per-session offsets are measured rather than assumed.
Camera frames carry host arrival times, and the constant per-camera latency (USB transport plus exposure) is estimated once per rig build by cross-correlating a moving target seen by the camera and the LiDAR, with a resolution of about $\pm$\LatencyResMs\,\si{\milli\second}, and stored in the rig file.
That measurement had not been made before the earlier fused-cloud scenes session, so its camera timestamps are exported uncorrected and the association residual is recorded per frame instead.
Rolling-shutter line delay is not compensated.

\subsection{Calibration}
Intrinsics are estimated with OpenCV checkerboard calibration against an acceptance criterion of reprojection RMS below \IntrAcceptPx\,\si{px} at \IntrAcceptCoverage\,\% image coverage, and camera-to-LiDAR extrinsics use targetless direct visual-LiDAR calibration with normalized-information-distance refinement~\cite{koide2023calib}, seeded from the measured mount positions.
Calibration is re-run before every recording session, the session-frozen rig file is what the exporter reads (the release section of the main paper), and every exported calibration entry records whether it is measured or nominal.
In the earlier fused-cloud scenes every intrinsic is factory or nominal and no distortion model is shipped, which the release metadata states.

\subsection{Ego Pose and Ground Truth}
\label{supp:egopose}
Six-DoF ego poses are produced offline in a fixed order of preference.
The first tier fuses PPK positions with INS attitude in a local ENU frame shared by all sessions, with the antenna lever arm applied at full attitude.
The second tier anchors LiDAR-inertial odometry to the available PPK epochs by Umeyama alignment and records the residual, and the third tier is odometry alone, flagged as having no global reference.
As an independent check we run LiDAR-inertial odometry on every sequence, with FAST-LIO2~\cite{xu2022fastlio2} by default and GLIM~\cite{koide2024glim}, FAST-LIVO2~\cite{zheng2025fastlivo2} and its multi-camera extension Omni-LIVO~\cite{cao2026omnilivo} where the cameras are calibrated, and report the disagreement as ATE after alignment beside the PPK fix status, so that users know the trajectory quality of each sequence rather than assuming centimetre accuracy throughout.
On the earlier fused-cloud scenes session the antenna lever arm had not been surveyed and no PPK epoch fixed, so its poses sit in the second tier and no LIO disagreement is reported yet.

\subsection{Design Choices Behind the Rig}
The Mid-360 was chosen over a spinning unit for cost and for its \SI{360}{\degree} by \SI{59}{\degree} field of view, which covers the deck and the near road surface from a single mast.
Its non-repetitive scan pattern fills the field of view over several sweeps, which is why ingestion accumulates five ego-motion-compensated sweeps before ground removal.
The six cameras were chosen as the cheapest way to close the ring: two ZED~2i stereo pairs give calibrated stereo depth front and back, the two RealSense units give depth on the flanks, and the two Logitech webcams fill the front diagonals.
The price is that every camera is rolling shutter and free running, which the main paper states as a limitation and which the time synchronization procedure above measures rather than hides.
The SBG Ellipse-D replaced the plan of a second u-blox receiver because one F9P cannot serve both timing and PPK, as measured in the receiver conflict described in the main paper, and its dual-antenna heading removes the magnetometer from the attitude chain.
The VectorNav VN-200 stays on the rig as a fallback attitude and RTK source, recorded by default since the second collection day.

\section{Collection Protocol and Sessions}
\label{supp:collection}
\begin{table}[t]
  \centering\footnotesize
  \caption{Every recorded session, colour coded as in the main paper's route map. Distance and duration from the post-processed trajectory, and the share of epochs whose PPK solution was fixed, float or code differential. \srcnote}
  \label{stab:sessions}
  \IfFileExists{table_sessions.tex}{\begin{adjustbox}{max width=\columnwidth}\begin{tabular}{@{}lrrrrr@{}}
\toprule
Session & Distance (km) & Duration (min) & Fixed (\%) & Float (\%) & DGPS (\%) \\
\midrule
S1 & 1.4 & 6 & 6 & 85 & 9 \\
S2 & 19.5 & 92 & 1 & 10 & 89 \\
S3 & 8.9 & 50 & 0 & 0 & 100 \\
S4 & 12.5 & 50 & 2 & 91 & 7 \\
S5 & 2.0 & 15 & 4 & 94 & 2 \\
S6 & 13.7 & 67 & 2 & 98 & 0 \\
S7 & 13.5 & 51 & 11 & 86 & 3 \\
\midrule
Total & 71.5 & 331 & & & \\
\bottomrule
\end{tabular}
\end{adjustbox}}{\tbd{run the harness}}
\end{table}
The campaign plan drives one district per day in sessions of about five hours split into three recording legs, with the recorder stopped between legs so that legs remain separable routes for the route-disjoint split.
The planned districts were chosen to span the density spectrum rather than to maximize distance: the central business district around Motijheel, Gulistan and Paltan with the Hatirjheel loop for loop closures, the diplomatic zone of Gulshan, Banani and Baridhara, the western commercial district of Dhanmondi, New Market and Mohammadpur, the dense residential blocks of Mirpur, Kazipara and Shewrapara, and the Airport Road arterial toward Uttara.
Recording produces about \RecGbPerHour\,\si{GB} per hour raw and \RecGbPerHourCompressed\,\si{GB} per hour after reversible compression, in \ChunkSeconds\,\si{\second} chunks sealed with a content manifest, and two pre-flight gates refuse to record on battery power or while GNSS horizontal accuracy exceeds \GnssGateM\,\si{\metre}.

Table~\ref{stab:sessions} lists the \NSessionsRecorded{} sessions recorded so far on \NRecordingDays{} days, \RecordedKm\,\si{\kilo\metre} in \RecordedHours{} hours, whose trajectories the main paper maps.
Trajectories were post-processed with RTKLIB in kinematic mode with forward and backward combination against the project's own base station, a u-blox ZED-F9P logging at \SI{1}{\hertz}, using the SBG receiver's raw observables reassembled from the recorded serial stream.
On the first collection day the base station's L1 carrier phase is absent from its log for the whole recording window, so every phase double difference fails and the solution falls back to code differential at metre level, which is why the earlier fused-cloud scenes's metadata reports no fixed epochs.
On the second day the base is healthy and the sessions are between 86 and 98 percent float with up to 11 percent fixed.
Code-differential positions are adequate for the city-scale route map but not for the first pose tier, and the earlier fused-cloud scenes's poses therefore sit in the second tier.

\section{Taxonomy, Label Tiers and Statistics}
\label{supp:taxonomy}
\begin{table}[t]
  \centering\scriptsize
  \caption{The full \dsense{} taxonomy. $^\star$Headline long-tail classes. The earlier fused-cloud scenes annotates the \BenchNClasses{} classes the detector arms can name. \srcnote}
  \label{stab:taxonomy}
  \IfFileExists{table_taxonomy.tex}{\begin{tabular}{@{}llp{5.1cm}@{}}
\toprule
Group & \# & Classes \\
\midrule
Motorised 4-wheel & 3 & Car, bus, truck \\
Indigenous 3-wheel & 1 & \textbf{CNG auto-rickshaw}$^\star$ \\
Human-powered & 2 & \textbf{Cycle rickshaw}$^\star$, bicycle \\
Motorised 2-wheel & 1 & Motorcycle \\
Vulnerable users & 1 & Pedestrian \\
\midrule
Total & 8 & $^\star$headline long-tail classes \\
\bottomrule
\end{tabular}
}{\tbd{run the harness}}
\end{table}
\begin{table}[t]
  \centering\footnotesize
  \caption{Released instances and boxes per class in the earlier fused-cloud scenes. $^\star$Headline classes. \srcnote}
  \label{stab:classfreq}
  \IfFileExists{table_classfreq.tex}{\begin{tabular}{@{}lrrr@{}}
\toprule
Class & Instances & Boxes & Share (\%) \\
\midrule
Pedestrian & 14,557 & 50,759 & 49.5 \\
Car & 7,357 & 29,776 & 25.0 \\
Motorcycle & 2,610 & 9,048 & 8.9 \\
CNG auto-rickshaw$^\star$ & 1,220 & 6,807 & 4.2 \\
Cycle rickshaw$^\star$ & 1,129 & 5,861 & 3.8 \\
Bicycle & 1,058 & 2,347 & 3.6 \\
Truck & 846 & 3,151 & 2.9 \\
Bus & 602 & 2,064 & 2.0 \\
\midrule
Total & 29,379 & 109,813 & 100.0 \\
\bottomrule
\end{tabular}
}{\tbd{run the harness}}
\end{table}
Table~\ref{stab:taxonomy} lists the \NClasses{} classes of the full taxonomy, grouped by propulsion and steering.
The groups are motorised four-wheelers, indigenous three-wheelers, human-powered vehicles, motorised two-wheelers and vulnerable road users, and the grouping follows how a vehicle moves in mixed traffic rather than its size.
The earlier fused-cloud scenes annotates the \BenchNClasses{} classes that the detector arms can name.
Battery rickshaws (easy-bikes) are labelled as cycle rickshaws because arm B's rickshaw class does not separate the two, and the covered van, microbus, tempo, human hauler, pushcart and animal are outside the released taxonomy.

Labels carry one of three tiers.
Verified keyframes carry human-corrected 3D boxes, track identities and instance masks and form the val and test splits once the human pass completes.
Pipeline keyframes carry auto-generated labels with a per-box confidence tier, admitted at the auto-accept tier of the quality gate.
Unverified keyframes carry pipeline labels that no person has reviewed, which in the earlier fused-cloud scenes is every keyframe.
Boxes that the gate flagged or rejected are not deleted but shipped in a side table with the reason, so that a future human pass can recover them, and interpolated boxes that fail the return floor at their own keyframe go to the same table.
Table~\ref{stab:classfreq} gives the released counts per class.

\section{Annotation Pipeline Stage Reference}
\label{supp:annotation}
\begin{table*}[t]
  \centering\scriptsize
  \caption{Pipeline parameters as implemented, with the source file of each value. Values are read from the pipeline repository at the revision named in the release metadata.}
  \label{stab:params}
  \setlength{\tabcolsep}{4pt}
  \begin{tabular}{@{}llll@{}}
    \toprule
    Stage & Parameter & Value & Source \\
    \midrule
    Probe & sweep window / coverage / max gap & \SI{0.5}{\second} / \SI{80}{\percent} of 5 sweeps / 3 periods & \texttt{stage0\_data\_probe/probe.py} \\
    Probe & point-cloud size band / JPEG byte band & 10k--300k points / 20~kB--4~MB & \texttt{common/schemas.py} \\
    Ingestion & RANSAC sectors / iterations / inlier threshold & 8 / 200 / \SI{0.10}{\metre} & \texttt{stage1\_ingestion/ingest.py} \\
    Ingestion & ground band / tilt reject / height disagreement & \SI{0.30}{\metre} / \SI{15}{\degree} / \SI{0.25}{\metre} & \texttt{stage1\_ingestion/ingest.py} \\
    Ingestion & range cap / height cap / accumulated sweeps & \SI{50}{\metre} / \SI{4}{\metre} / 5 & \texttt{stage1\_ingestion/ingest.py} \\
    Proposals & input size / confidence floor / NMS same, cross & 1600~px / 0.40 / 0.65, 0.90 & \texttt{stage3\_proposals/proposals.py} \\
    Proposals & min box side / max proposals per image & 4~px / 300 & \texttt{stage3\_proposals/proposals.py} \\
    Proposals & arm A model, revision & YOLO11x COCO, v8.3.0 & \texttt{scripts/run\_stages.sh} \\
    Proposals & arm B model, revision & YOLO11x RSUD20K fine-tune, armb-r1280-4 & \texttt{scripts/run\_stages.sh} \\
    Merge & suppression IoU / bicycle veto floor / containment & 0.5 / 0.40 / 0.6 & \texttt{stage3\_merge/merge.py} \\
    Propagation & window / match IoU / miss tolerance / decay & 16 frames / 0.30 / 2 keyframes / 0.7 per hop & \texttt{stage3b\_track2d/track2d.py} \\
    Propagation & area ratio / frame fraction / tracks per camera & 2.5 / 0.25 / 64 & \texttt{stage3b\_track2d/track2d.py} \\
    Masks & model / min mask / IoA threshold & SAM~3.1 / 16~px / 0.5 & \texttt{stage4\_masks/masks.py} \\
    Lift & min depth / min points recorded & \SI{1e-3}{\metre} / 5 & \texttt{stage5\_lift/lift.py} \\
    Cluster & DBSCAN min samples / eps scale / eps fallback & 3 / 0.6 / \SI{0.75}{\metre} & \texttt{stage6\_cluster/cluster.py} \\
    Cluster & L-shape step / refinements / near-square ratio / min yaw points & \SI{1}{\degree} / 3$\times$10 / 0.80 / 10 & \texttt{stage6\_cluster/cluster.py} \\
    Track & IoU gate / confirm hits / max misses & 0.05 / 1 / 2 & \texttt{stage7\_track/track.py} \\
    Track & process noise pos, vel, yaw & \SI{0.5}{\metre}, \SI{2}{\metre\per\second}, \SI{5}{\degree} per second & \texttt{stage7\_track/track.py} \\
    Track & measurement noise pos, yaw / ICP points, iterations, tolerance & \SI{0.3}{\metre}, \SI{10}{\degree} / 15, 25, \SI{1e-4}{\metre} & \texttt{stage7\_track/track.py} \\
    Inflate & trigger points / max growth ratio, metres / sigma clamp & 15 / 3.0, \SI{3}{\metre} / 0 & \texttt{stage8\_inflate/inflate.py} \\
    Gate & confidence / returns / spatial multiple & 0.5 / 5 / 2.0 & \texttt{stage9\_qa/gate.py} \\
    Release & stitch max gap / base gate / slack / size ratio & \StitchMaxGap{} / \StitchBaseGateM\,\si{\metre} / \StitchSlackM\,\si{\metre} / \StitchSizeRatio & \texttt{conrelease.yaml} \\
    Release & interpolated point floor / moving threshold / max $\Delta t$ & \StitchMinPts{} / \MovingSpeedMps\,\si{\metre\per\second} / \AttrMaxDtS\,\si{\second} & \texttt{conrelease.yaml} \\
    Release & density radius, quantiles / luma edges / double fraction & \DensityRadiusM\,\si{\metre}, \DensityQuantiles{}\,\% / \IllumEdges{} / \DoubleFractionPct\,\% & \texttt{conrelease.yaml} \\
    \bottomrule
  \end{tabular}
\end{table*}

Annotating 3D boxes by hand at nuScenes scale is beyond a university group's budget.
\dsense{} labels are therefore produced by a pipeline that runs from a raw sensor export to gated 3D boxes, tracks and review tasks without a person in the loop, replacing human judgement with named predicates, refusals and provenance (Fig.~2 of the main paper).
People enter afterwards, at defined points, to verify a stratified subset and to correct the published review tasks, so that the released labels are semiautomatic rather than either hand-drawn or blindly machine-generated.
This section describes each stage as implemented, with the thresholds that the code uses.

\subsection{Three State Exits and Provenance}
Every stage exits in one of three states.
A clean exit writes a success marker, a degraded exit writes a marker that names its causes, and a refusal writes nothing, so that an absent marker is unconditionally a refusal and a crash can never pass as a quality flag.
Each stage also writes a manifest binding its output to a SHA-256 fingerprint of the substrate, the resolved configuration with a provenance string per key, and pinned model revisions, and a downstream stage refuses a fingerprint mismatch or an unaccepted degraded upstream.
The orchestrator cross-examines each exit code against the marker on disk and chooses which proposal tree the mask stage consumes by marker, freshness and scene coverage.
Per box, the record carries which arm proposed it, what suppressed or protected it, the return-count basis, the measured and the inflated geometry, and the gate vector.
On the earlier fused-cloud scenesd scene, \NDegraded{} of \NMarkers{} stage markers are degraded, each naming its cause, which we regard as the mechanism working rather than failing.

\subsection{Stages}
\subsubsection{Substrate Probe}
A probe admits scenes through six named predicates and excludes a failing scene with the predicate named rather than repairing it.
The predicates are version match, channel completeness, file resolution, file parsing within byte and point-count bands, sweep coverage, and token-graph closure.
Sweep coverage requires at least \SI{80}{\percent} of the five expected sweeps in the \SI{0.5}{\second} window before each keyframe, with no gap above three sweep periods.
The resulting scene list is the only input every later stage reads.

\subsubsection{Ingestion and Ground Removal}
Ingestion applies the sensor-to-ego calibration once, accumulates five ego-motion-compensated sweeps through the sensor, ego, global and anchor-ego frames, and fuses the stereo depth streams into the single sweep.
Ground is removed by a sector-wise seeded RANSAC plane fit over eight azimuth sectors (200 iterations, \SI{0.10}{\metre} inlier threshold), and a sector whose plane tilts more than \SI{15}{\degree} or disagrees in height by more than \SI{0.25}{\metre} with its neighbours falls back to a whole-cloud reference plane.
Points within \SI{0.30}{\metre} of the fitted ground, beyond \SI{50}{\metre} range or above \SI{4}{\metre} are dropped from the single sweep only, and the accumulated cloud is kept unfiltered for clustering context.
A watchdog aborts a keyframe after \SI{300}{\second} and the stage after three such strikes.

\subsubsection{Two Arm 2D Proposals}
No COCO-vocabulary detector can name a rickshaw, and open-vocabulary detectors prompted with the class phrases cross-fire between phrases that share a head noun.
Two arms therefore run on every camera frame with the same adapter.
Arm~A is YOLO11x~\cite{khanam2024yolo11} with its COCO vocabulary, of which six classes map onto the taxonomy.
Arm~B is the same architecture fine-tuned on RSUD20K~\cite{zunair2024rsud20k}, a Bangladeshi road-scene dataset of \ArmBTrainImages{} training images, from which only the rickshaw and CNG auto-rickshaw outputs are kept, because person, car and motorcycle are trained alongside as boundary classes and never emitted.
Both arms run at an input size of 1600~px with a per-class confidence floor of 0.40, followed by greedy non-maximum suppression at IoU 0.65 within a class and 0.90 across classes, dropping boxes with a side under 4~px and capping 300 proposals per image.

\subsubsection{Vocabulary Authority Merge}
The arms are merged by a fixed authority table, not by score.
Where an arm-B rickshaw-family box overlaps an arm-A car, truck, bus, motorcycle or bicycle with $\mathrm{IoU}>0.5$, the arm-A box is suppressed, because arm~A is confident on its wrong answers, while pedestrians from both arms are kept.
One veto applies, namely that an arm-A bicycle with confidence at least 0.40 is never suppressed and instead vetoes every arm-B box overlapping it, added after the pilot showed arm~B over-calling rickshaws on plain bicycles.
A rider box is absorbed by the rickshaw that contains it when
\begin{equation}
  \mathrm{cont}(a, b) = \frac{|a \cap b|}{|a|} \ge 0.6,
\end{equation}
where $a$ is a surviving arm-A bicycle or motorcycle and $b$ a surviving arm-B rickshaw or auto-rickshaw, so the wheels of a rickshaw are not labeled twice.
Nothing is deleted, and a suppressed, vetoed or absorbed box moves to a ledger on the row that beat it.

\subsubsection{Blind Check and Temporal Propagation}
Two optional stages bracket the merge.
Before it, a blind check re-classifies each proposal crop with a local vision-language model that never sees the current label, rewrites the label where the model confidently disagrees, skips crops below a minimum side length and the rider classes it was measured to flip, records a verdict per box and a confusion table per run, and leaves geometry, scores and row order untouched, so that an arm's systematic confusion is corrected before the authority table acts on it.
After it, the SAM~3.1 video tracker~\cite{carion2025sam3} propagates the merged boxes through the intervening \SI{12}{\hertz} camera frames in windows of at most 16 frames, associating tracks to detections by Hungarian assignment at $\mathrm{IoU}\ge0.30$ within a class, killing a track after two missed keyframes, and decaying a recovered box's score as $0.7^{h}$ over $h$ propagation hops, with a leak guard that rejects recovered boxes above 2.5 times their seed area or a quarter of the frame. Every propagated box lands at a keyframe with its provenance and a 2D identity.
The propagation stage ran on the released scenes, and the blind check did not.

\subsubsection{Masks and Cross Camera Suppression}
Each surviving box prompts SAM~3.1 for one instance mask, and masks under 16~px are dropped.
Because adjacent cameras overlap, the same object is often masked twice, so masks are de-duplicated in the ego frame rather than in pixels.
Each mask's tight box is back-projected through the camera intrinsics and extrinsics to an azimuth-by-elevation footprint, and for two same-class candidates $a$ and $b$ ordered by score and then by a fixed camera priority, $b$ is suppressed when
\begin{equation}
  \mathrm{IoA}(a,b) = \frac{\Delta\mathrm{az}(a\cap b)\,\Delta\mathrm{el}(a\cap b)}{\Delta\mathrm{az}(a)\,\Delta\mathrm{el}(a)} > 0.5 .
\end{equation}

\subsubsection{Painting LiDAR Points}
Masks are transferred to the single sweep by projecting every LiDAR point into every camera through four hops, from the ego frame at the LiDAR time to the global frame, back into the ego frame at that camera's exposure time, into the camera frame and through the pinhole $\mathbf{u} = \pi(\mathbf{K}\,\mathbf{p}_{\mathrm{cam}})$ without a distortion model, so that the free-running cameras are compensated for ego motion between the two timestamps~\cite{vora2020pointpainting}.
A point in front of the camera and inside the image is assigned to a mask by a raster look-up, and a point seen by several cameras is awarded to the camera whose optical axis it lies closest to, with a fixed camera priority as tie-break.
The point count per instance is recorded here on an asserted single-sweep, ground-filtered basis and is not filtered until the gate.

\subsubsection{Clustering and Box Fitting}
Per instance, the painted points are clustered in the bird's-eye view by DBSCAN with a minimum of three points and a class-conditional radius
\begin{equation}
  \varepsilon_c = 0.6\,\overline{\sqrt{w_c^{2}+l_c^{2}}},
\end{equation}
the scaled mean diagonal of that class's prior footprint, falling back to \SI{0.75}{\metre} without a prior.
The largest cluster is fitted with an L-shape by a brute-force search over the heading $\theta\in[0,\pi/2)$ that maximizes the closeness criterion of Zhang et al.,
\begin{equation}
  S(\theta) = \sum_{i} \frac{1}{\max\bigl(\min(d^{u}_{i}, d^{v}_{i}),\,0.01\bigr)},
\end{equation}
where $d^{u}_i$ and $d^{v}_i$ are the distances of point $i$ to the nearer edge along the rotated axes, searched at \SI{1}{\degree} and refined three times by a factor of ten.
The longer extent becomes the heading axis, a box whose width-to-length ratio exceeds 0.80 or that has fewer than ten points is flagged as yaw ambiguous, and a cluster whose measured extent exceeds twice the class prior is re-clustered on the points within \SI{5}{\metre} of its nearest return so that a fragment of background is not fitted as part of the object.

\subsubsection{Tracking}
Boxes are linked across keyframes by a Kalman filter over the state $(x, y, z, v_x, v_y, v_z, \psi)$ with a constant-velocity, constant-yaw motion model, position and yaw measurements with standard deviations of \SI{0.3}{\metre} and \SI{10}{\degree}, and process noise of \SI{0.5}{\metre}, \SI{2}{\metre\per\second} and \SI{5}{\degree} per second~\cite{weng2020ab3dmot}.
Association is Hungarian over the cost $-\mathrm{IoU}_{\mathrm{BEV}}$, admitting only same-class pairs with $\mathrm{IoU}_{\mathrm{BEV}}\ge0.05$, and where a re-identification embedding is trusted the score is multiplied by its cosine similarity.
When both clusters hold at least 15 points, point-to-point ICP between adjacent keyframes in the global frame supplies a velocity measurement that updates, rather than overwrites, the filter.
A track is confirmed at its first hit and dies after two consecutive misses, and yaw consistency along the velocity vector is checked above \SI{1}{\metre\per\second}.

\subsubsection{Inflation Toward Class Priors}
A box fitted to few returns is systematically too small, so a box with $n$ points below 15 is grown toward the class-mean dimensions $\mu_c$ by a clamped one-sided blend,
\begin{align}
  \alpha &= \min\!\Bigl(1, \max\!\bigl(0, 1 - \tfrac{n}{15}\bigr)\Bigr), \label{seq:inflate}\\
  d^{\star} &= \max\bigl(d,\ \min(d + \alpha(\mu_c - d),\ \mu_c,\ 3d,\ d + 3)\bigr),
\end{align}
for each measured dimension $d$ in metres, so a box never shrinks, never exceeds the class mean, and never grows more than threefold or by more than \SI{3}{\metre}.
Growth anchors the sensor-facing face and the box top, so only the far face and the underside move, a guard refuses any growth toward the sensor, and the measured box is retained beside the shipped one.

\subsubsection{Quality Gate}
The gate evaluates three predicates on the pre-inflation box,
\begin{equation}
  g_1 = [\,c \ge 0.5\,],\;
  g_2 = [\,n \ge 5\,],\;
  g_3 = [\,s \le 2 s_{\mu} \wedge \ell \le 2 \ell_{\mu}\,],
  \label{seq:gate}
\end{equation}
where $c$ is detector confidence, $n$ the single-sweep return count, and $s,\ell$ and $s_\mu,\ell_\mu$ the shorter and longer footprint sides of the box and of the class prior.
A box is rejected if $g_2$ or $g_3$ fails, flagged if $g_1$ fails or no prior exists for its class, and otherwise admitted at the auto-accept tier.
A tier never upgrades the record's source, which stays \emph{pipeline}.

\subsubsection{Release Writer}
Gated records are converted into nuScenes tables in a fixed order of stitching, tier partition, attribute assignment, stratification and table writing, with the global frame taken from the keyframe ego pose and instance tokens from track identities.
Because the tracker confirmed only \RelTracksConfirmed{} tracks over the released scenes, identities are stitched after the gate across gaps of up to \StitchMaxGap{} keyframes by Hungarian assignment on the normalized distance $d/\gamma(k)$ with
\begin{equation}
  \gamma(k) = \StitchBaseGateM + \StitchSlackM\,(k-1)\ \si{\metre},\qquad k = 1,\dots,\StitchMaxGap,
\end{equation}
for a gap of $k$ keyframes, where $d$ is the smaller of the velocity-extrapolated and the raw centre distances, and a join also requires a footprint size ratio at most \StitchSizeRatio.
Gaps inside a chain are filled by linear interpolation of position and size and spherical interpolation of orientation, and an interpolated box ships only if at least \StitchMinPts{} returns fall inside it.
Motion attributes come from the symmetric finite-difference velocity along the chain, valid when the neighbouring keyframes are at most \AttrMaxDtS\,\si{\second} apart at a chain end and twice that inside it, with a threshold of \MovingSpeedMps\,\si{\metre\per\second} between moving and stopped.
The writer computes the density and illumination strata of the benchmark section of the main paper, refuses the export on any category it cannot map onto the taxonomy, and publishes review tasks to CVAT~\cite{cvat2020} additively under a run tag, so a new run lands beside an earlier review instead of destroying it.

\begin{table}[t]
  \centering\footnotesize
  \caption{Wall-clock time per stage over the five released scenes on one GPU. The opt-in propagation arm dominates and is the first candidate to drop at scale. \srcnote}
  \label{stab:runtime}
  \IfFileExists{table_stage_runtime.tex}{\begin{tabular}{@{}lrr@{}}
\toprule
Stage & Hours & Share (\%) \\
\midrule
Ingestion & 0.2 & 0 \\
Arm A proposals & 0.4 & 1 \\
Arm B proposals & 0.3 & 1 \\
Propagation (opt-in) & 25.7 & 67 \\
SAM 3.1 masks & 4.3 & 11 \\
Painting & 0.2 & 1 \\
Clustering and fit & 3.4 & 9 \\
Tracking & 4.1 & 11 \\
Inflation & 0.0 & 0 \\
Gate & 0.0 & 0 \\
\midrule
Total & 38.6 & 100 \\
\bottomrule
\end{tabular}
}{\tbd{run the harness}}
\end{table}
\subsection{Reach, Computational Cost and the Human Boundary}
\label{supp:humanboundary}
With both arms the detectors name \NReachable{} of the \NClasses{} classes.
The covered van and microbus have no detector path and are outside the released taxonomy.
On \RelNChunksDone{} of the \SessNScenes{} scenes of the first session (\RelNSamples{} keyframes) the chain ran to the release writer for the first time on Dhaka data.
On the first scene (\PilotKeyframes{} keyframes) arm~A produced \NArmA{} proposals and arm~B \NArmB, the merge emitted \NMerged{} (\NSuppressed{} suppressed, \NVetoed{} vetoed), and \NMasksKept{} masks painted \NPainted{} LiDAR points into \NBoxes{} boxes.
Across the five scenes the gate admitted \RelNAnnotations{} boxes on \RelNInstances{} stitched instances and withheld \RelNExcluded, and the whole chain took \RelChainHours\,\si{\hour} on one GPU, of which \RelPropagationHours\,\si{\hour} was the \SI{12}{\hertz} propagation arm, so that arm is the first candidate to drop at scale.

Humans enter at four points only.
Annotators review the published tasks in CVAT, where they correct, delete or add boxes and attest each frame so coverage is auditable.
Domain experts author the class dimension priors and tune the per-class thresholds.
Finally, a stratified \DoubleFractionPct\,\% of keyframes, allocated proportionally over the density-by-illumination cells with at least one keyframe per non-empty cell, is fully annotated by humans independently of the pipeline and used to validate the released labels, from which we report the pipeline's precision and recall against the human boxes and the inter-annotator agreement (3D IoU and Cohen's $\kappa$~\cite{cohen1960kappa}).
Labels descended from arm~B inherit the CC~BY-NC~4.0 terms of RSUD20K, which the release records per box.
At submission the human pass on the earlier fused-cloud scenes was published to CVAT but had not completed, so every released box is pipeline output and the validation statistics are not yet available.

\section{Release Format and Known Deviations}
\label{supp:release}
\dsense{} is released in the nuScenes table format so that existing tooling loads it unchanged.
Every LiDAR frame is a keyframe, camera frames between keyframes are sweeps, camera extrinsics are stored in the optical frame with the body-frame rotation kept in an extra field, one ego pose is interpolated per sensor record, and a stub map record is present.
The stereo depth clouds of the two ZED cameras are shipped either fused into the LiDAR file with ring identifiers 100 and 101, as in the earlier fused-cloud scenes, or as their own channel in the global frame, as in later exports, and the release metadata states which.
Each annotation row carries the standard nuScenes fields plus a set of prefixed fields that record its provenance, namely the record token, the source (pipeline or human), the tier and its basis, whether the row is interpolated, the stitched chain identifier and the pre-stitch track identifier, the basis on which returns were counted, the visibility basis, the chain velocity, the attribute basis, the verifier, and an uncertainty flag with its reason.
The return count is measured on the single ground-filtered sweep before inflation, and in exports that fuse stereo points it counts LiDAR and stereo returns together, which the release note states.

The earlier fused-cloud scenes's metadata lists its deviations from the format's assumptions so that users do not have to discover them: no exposure or gain metadata is available for the webcams, intrinsics are nominal with no distortion model, the left and right RealSense mapping is to be verified, the session covers a single area on a single afternoon so the route identifier is provisional, the inertial data ship without extrinsics, and face and plate blurring is applied to the released images after annotation so that boxes drawn from image evidence saw the original pixels.
Each sequence also ships a sensor-quality record with the time-sync report, calibration residuals, the PPK fix ratio, the camera latency values used, and a per-entry flag stating whether each calibration value was measured or nominal.
The data are released under CC~BY-NC-SA~4.0, the capture stack is GPL-3.0, and labels descended from the fine-tuned detector arm inherit the CC~BY-NC~4.0 terms of its training set and are marked per box.

\section{Benchmark Protocol Details}
\label{supp:benchmark}
\begin{figure*}[t]
  \centering
  \IfFileExists{fig_bench.pdf}{\includegraphics[width=0.62\textwidth]{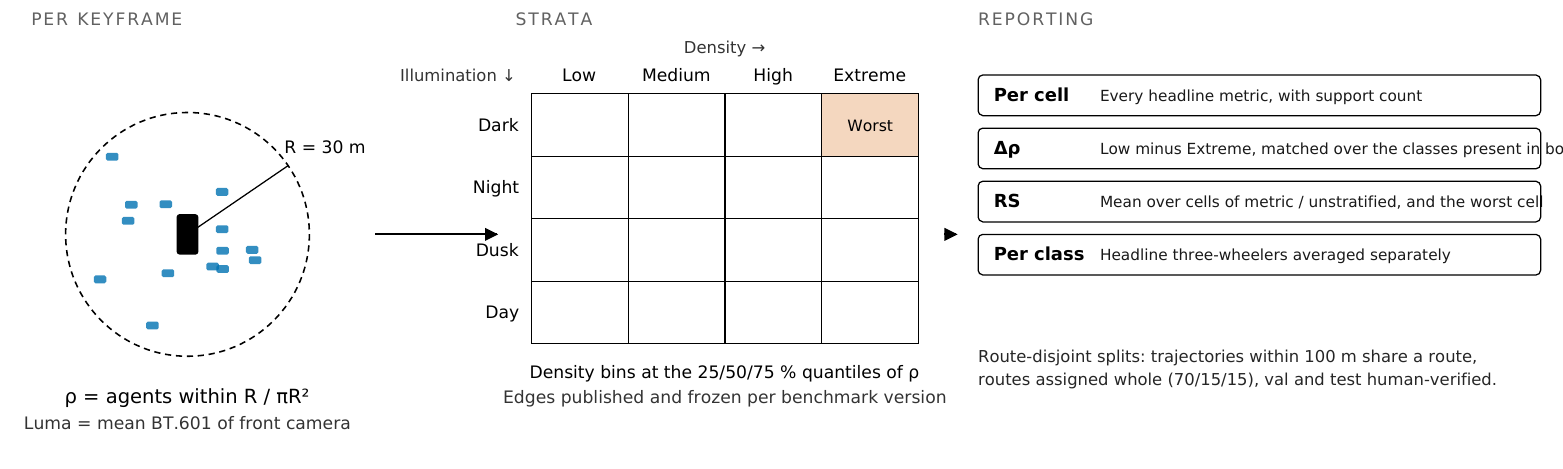}}{\figtbd{run make\_figs.py}}
  \caption{The benchmark protocol. Each keyframe is binned by road-user density within \DensityRadiusM\,\si{\metre} and by front-camera brightness. Every metric is reported per cell with a matched-class degradation and a Robustness Score beside its worst cell, and splits are route-disjoint. \srcnote}
  \label{sfig:bench}
\end{figure*}
The evaluation code, \dbench, wraps the official nuScenes and KITTI metric kernels without modification and adds a stratified reporting layer (Fig.~\ref{sfig:bench}).
Its detection kernel is gated to agree with the nuScenes devkit to $10^{-6}$ on nuScenes data, and every result file records the SHA-256 of the benchmark definition that produced it.

\subsection{Tasks and Metrics}

\textbf{3D Object Detection:} A prediction matches a ground-truth box of the same class when their bird's-eye-view centres lie within a threshold $d\in\{0.5,1,2,4\}\,\si{\metre}$. Average precision is the normalized area under the precision-recall curve above 10\,\% recall and precision, averaged over the four thresholds and the eight classes. The true positive errors are the translation error (mATE, m), the scale error (mASE, $1-\mathrm{IoU}$ after centre and yaw alignment), the orientation error (mAOE, rad), the velocity error (mAVE, m/s) and the attribute error (mAAE, $1-\mathrm{acc}$), each computed on matches at \SI{2}{\metre}, and the nuScenes detection score is $\mathrm{NDS} = \frac{1}{10}\bigl[5\,\mathrm{mAP} + \sum_{e}(1-\min(1,e))\bigr]$ over the five errors~\cite{caesar2020nuscenes}. Ground truth is filtered to the per-class range bands of the benchmark definition, to boxes with at least five points, and to the drivable zone.

\textbf{Point Segmentation:} Per-point labels are released on two layers, the eight object classes on the stereo cloud (\SegZedLabelledPct\,\% of its points labelled) and the road class on the LiDAR sweep (\SegLidarLabelledPct\,\% of its points). mIoU follows nuScenes lidarseg with an ignore index, is computed over the classes a method is eligible for (a detector cannot propose a stuff class, so road is excluded from a box-derived method's mean and recorded as ineligible rather than scored zero), and a method is scored on the cloud it consumes.

\textbf{Tracking:} Not part of this benchmark version. The released instance identities persist for a median of \TrackMedianFrames{} keyframes, so identity metrics would measure annotation churn rather than a tracker, and the task is withdrawn from the benchmark definition.

\textbf{Depth:} The stereo output is a sensor stream and is never used as ground truth. It is scored against the LiDAR sweep projected into the same camera through the same projection chain, so a residual is the two sensors disagreeing and not two implementations. Metrics are AbsRel, SqRel, RMSE, RMSE$_{\log}$ and $\delta<1.25^{k}$~\cite{eigen2014depth} over 0.1 to \SI{80}{\metre}, reported per illumination bin, and a sparse prediction is filled to its 8-pixel neighbourhood before scoring so that the sampling pattern is not what is measured.

\textbf{Occupancy:} Ground truth is a $\pm4$ keyframe accumulation with ego-motion compensation on \SI{0.4}{\metre} voxels over $x,y\in[-40,40]$\,m and $z\in[-1,5.4]$\,m. The window is short on purpose, since a long one turns moving traffic into permanent walls. Voxel mIoU is reported beside RayIoU~\cite{liu2024sparseocc} at 1, 2 and \SI{4}{\metre}, which compares first-hit depths along rays and therefore separates sampling density from surface accuracy.

\textbf{Odometry and Localization:} Following the KITTI odometry benchmark~\cite{geiger2012kitti}, the relative translation error $t_{\mathrm{rel}}$ in percent and the relative rotation error $r_{\mathrm{rel}}$ in degrees per metre are averaged over all sub-sequences of length 100, 200, \dots, \SI{800}{\metre}. The absolute trajectory error after SE(3) Umeyama alignment and the relative pose error at \SI{1}{\second} and at \SI{10}{\metre} and \SI{100}{\metre}~\cite{sturm2012benchmark} are reported beside them. Estimates are associated with the \SI{1}{\hertz} reference within \SI{50}{\milli\second}. The reference's fixed-solution ratio and position sigma are attached to every result because an error at or below them is a property of the reference. An estimator that consumes the PPK solution is scored as a localization system with its along-track, cross-track and vertical RMSE and bias, and with the pose success rate at \SI{0.5}{\metre} and \SI{2}{\degree}.

\subsection{Density Stratified Evaluation}
\label{supp:density}
For each keyframe $t$ we compute the road-user density
\begin{equation}
  \dens_t = \frac{N_t(R)}{\pi R^{2}},\qquad
  N_t(R) = \bigl|\{\, i : \lVert \mathbf{p}_{i,t} - \mathbf{p}^{\mathrm{ego}}_{t} \rVert_2 \le R \,\}\bigr|,
  \label{seq:density}
\end{equation}
where $\mathbf{p}_{i,t}\in\mathbb{R}^2$ is the bird's-eye-view centre of annotated agent $i$, $\mathbf{p}^{\mathrm{ego}}_t$ is the ego position and $R=\DensityRadiusM\,\si{\metre}$.
Keyframes are binned into Low, Medium, High and Extreme at the \DensityQuantiles{}\,\% quantiles $q_{1}<q_{2}<q_{3}$ of the dataset's own $\dens$ distribution, so that bin $b(t)=1+\sum_{k}[\dens_t > q_k]$, and the edges are published and frozen with the benchmark version.
Every metric $m$ is reported per bin together with the degradation
\begin{align}
  \Delta_\dens &= m(\mathrm{Low}) - m(\mathrm{Extreme}),\\
  \Delta_\dens^{\mathrm{matched}} &= \frac{1}{|\mathcal{C}_\cap|}\sum_{c\in\mathcal{C}_\cap}\bigl[m_c(\mathrm{Low}) - m_c(\mathrm{Extreme})\bigr],
\end{align}
where $\mathcal{C}_\cap$ is the set of classes with ground truth in both bins.
The matched form is reported because class composition shifts with density, and on the nuScenes-mini development fixture the two forms differ in sign, which is a property of the protocol and not a Dhaka result.
A stratum with fewer ground-truth boxes than the benchmark's reliability floor is flagged and its value is not quotable.

\subsection{Illumination Strata}
Each keyframe is binned by the mean BT.601 luma of its front camera image,
\begin{equation}
  \bar{Y}_t = \frac{1}{|\Omega_t|}\sum_{u\in\Omega_t}\bigl(0.299\,R_u + 0.587\,G_u + 0.114\,B_u\bigr),
\end{equation}
with $\Omega_t = \{u : \max(R_u,G_u,B_u) \le 250\}$ the set of unsaturated pixels, into dark, night, dusk and day at fixed edges of \IllumEdges{}.
The edges were calibrated on nuScenes-mini and are provisional until re-estimated on \dsense.
The measure is image brightness rather than exposure-compensated luminance, because the consumer cameras expose no gain metadata, so it characterizes what a camera-based method actually sees.

\subsection{Robustness Score}
\begin{equation}
  \rs = \frac{1}{|\mathcal{S}|}\sum_{s\in\mathcal{S}} \tilde m_s,\qquad \tilde m_s = \frac{m_s}{m_{\mathrm{all}}},\qquad
  s^{\star} = \arg\min_{s\in\mathcal{S}} \tilde m_s,
  \label{seq:rs}
\end{equation}
where $\mathcal{S}=\mathcal{D}\times\mathcal{I}$ is the set of non-empty density-by-illumination cells, $m_s$ the primary metric in cell $s$ and $m_{\mathrm{all}}$ its unstratified value.
\rs{} is always reported beside its worst cell $s^\star$ and $\tilde m_{s^\star}$, following the mean-resilience-rate construction of Robo3D~\cite{kong2023robo3d}.

\subsection{Headline Classes}
Per-class results are always reported, and the \BenchNHeadline{} rickshaw classes are averaged separately from the nuScenes-compatible classes.

\subsection{Splits and Evaluation Protocol}
\label{supp:splits}
Keyframe trajectories are clustered with a \RouteSepM\,\si{\metre} proximity union-find and whole clusters are assigned to train, val and test, so that no test location is seen in training~\cite{roberts2017cross}.
The target ratio is \SplitFractions{} by keyframes, balanced over density and illumination strata, and the human-verified subset constitutes val and test.
Neither released session yields such a split, so the baselines of Section~\ref{supp:experiments} use provisional splits that are labelled as such.
Test labels are withheld, and the evaluation server accepts the nuScenes submission format, enforces one submission per method per day, and requires disclosure of training data and pretrained weights.

\section{Experiments in Full}
\label{supp:experiments}
This section carries the complete results behind Section~VII of the main paper with standard deviations over seeds, together with the diagnostics that support its readings.
Evaluation sequences are lettered A to E for the earlier fused-cloud scenes, where A is the held-in validation split of the training sequence, and F to I for the four benchmarked scenes.
Every number is read by the analysis harness from a scored result file, and the provenance stamp on each table names the files.

\subsection{Experimental Setup}
\textbf{Scene Sets:} The earlier fused-cloud scenes ship one fused point cloud per keyframe, of which \CloudStereoSharePct\,\% of the points are stereo depth and \CloudMidSharePct\,\% Mid-360 returns, and boxes fitted to the visible surface, whose extents grow with range. The four benchmarked scenes ships the Mid-360 sweep of \MidPtsPerFrame{} points per keyframe and the stereo cloud of \ZedPtsPerFrame{} points as separate channels, and boxes with a class-prior length and fitted width and height. Attributes and track links are populated on \DayTwoLinkedPct\,\% of its boxes.

\textbf{Splits:} On the earlier fused-cloud scenes the detectors are trained on a strided held-in split of sequence A and evaluated on its validation part and on sequences B and C, which are held out whole. Pooled splits are block-random at 70/20/10 over whole blocks of keyframes within each sequence. No split is route-disjoint.

\textbf{Training:} OpenPCDet from nuScenes initializations, 40 epochs for PointPillars and CenterPoint and 20 for BEVFusion with an nuImages-pretrained Swin-T image backbone, three seeds each on one 96\,GB GPU. The taxonomy has eight classes since the annotation owner confirmed that no other class will carry ground truth. Every result file is stamped with the benchmark definition hash \texttt{\BenchConfigHash}, and the tables mirror the benchmark repository at commit \texttt{\BlackwellHead}.

\begin{table*}[t]
\caption{3D object detection on the held-in split of the earlier fused-cloud scenes with the full nuScenes metric set, mean and standard deviation over three seeds. mAP and NDS in percent, mATE in m, mASE as $1-\mathrm{IoU}$, mAOE in rad, mAVE in m/s, mAAE as $1-\mathrm{acc}$. Best per column in bold. \srcnote}
\label{stab:detfull}
\centering\scriptsize
\IfFileExists{table_det_full.tex}{\begin{tabular}{@{}llcrrrrrrr@{}}
\toprule
Method & Mod. & Train seq. & mAP & mATE & mASE & mAOE & mAVE & mAAE & NDS \\
\midrule
PointPillars~\cite{lang2019pointpillars} & L & 1 & 13.7$\pm$0.6 & 0.74$\pm$0.02 & \textbf{0.53$\pm$0.03} & 0.95$\pm$0.03 & 3.24$\pm$0.09 & 0.36$\pm$0.05 & 21.0$\pm$1.0 \\
CenterPoint~\cite{yin2021centerpoint} & L & 1 & 10.8$\pm$0.2 & 0.73$\pm$0.01 & 0.56$\pm$0.01 & 1.08$\pm$0.07 & 3.32$\pm$0.11 & 0.40$\pm$0.05 & 18.5$\pm$0.4 \\
BEVFusion~\cite{liu2023bevfusion} & L+C & 1 & \textbf{21.7$\pm$6.8} & 0.69$\pm$0.06 & 0.59$\pm$0.03 & 1.05$\pm$0.08 & \textbf{3.16$\pm$0.12} & 0.28$\pm$0.01 & \textbf{25.3$\pm$4.4} \\
\midrule
BEVFusion~\cite{liu2023bevfusion} & L+C & 5 & 17.1$\pm$0.4 & \textbf{0.66$\pm$0.02} & 0.56$\pm$0.00 & \textbf{0.87$\pm$0.02} & 3.92$\pm$0.10 & \textbf{0.26$\pm$0.01} & 25.0$\pm$0.2 \\
\bottomrule
\end{tabular}
}{\tbd{run the harness}}
\end{table*}
\begin{table*}[t]
\caption{3D object detection on the earlier fused-cloud scenes per evaluation sequence, mean and standard deviation over three seeds. Percent except mATE in m and mAOE in rad. \srcnote}
\label{stab:gen}
\centering\scriptsize
\IfFileExists{table_gen.tex}{\begin{tabular}{@{}lclrrrrr@{}}
\toprule
Method & Train seq. & Eval seq. & mAP & Rickshaw AP & mATE & mAOE & NDS \\
\midrule
PointPillars~\cite{lang2019pointpillars} & 1 & A (held-in) & 13.7$\pm$0.6 & 24.0$\pm$1.5 & 0.74$\pm$0.02 & 0.95$\pm$0.03 & 21.0$\pm$1.0 \\
PointPillars~\cite{lang2019pointpillars} & 1 & B & 0.2$\pm$0.0 & 0.0$\pm$0.0 & 0.92$\pm$0.00 & 1.11$\pm$0.01 & 4.9$\pm$0.7 \\
PointPillars~\cite{lang2019pointpillars} & 1 & C & 1.5$\pm$0.3 & 1.2$\pm$0.7 & 0.96$\pm$0.05 & 1.25$\pm$0.12 & 9.9$\pm$1.1 \\
\midrule
CenterPoint~\cite{yin2021centerpoint} & 1 & A (held-in) & 10.8$\pm$0.2 & 15.8$\pm$0.3 & 0.73$\pm$0.01 & 1.08$\pm$0.07 & 18.5$\pm$0.4 \\
CenterPoint~\cite{yin2021centerpoint} & 1 & B & 0.1$\pm$0.0 & 0.1$\pm$0.1 & 0.84$\pm$0.01 & 1.23$\pm$0.00 & 6.8$\pm$0.2 \\
CenterPoint~\cite{yin2021centerpoint} & 1 & C & 2.1$\pm$0.1 & 2.0$\pm$0.2 & 0.90$\pm$0.02 & 1.39$\pm$0.02 & 9.5$\pm$0.5 \\
\midrule
BEVFusion~\cite{liu2023bevfusion} & 1 & A (held-in) & 21.7$\pm$6.8 & 31.8$\pm$12.2 & 0.69$\pm$0.06 & 1.05$\pm$0.08 & 25.3$\pm$4.4 \\
BEVFusion~\cite{liu2023bevfusion} & 1 & B & 5.4$\pm$3.0 & 12.2$\pm$8.7 & 0.65$\pm$0.02 & 1.27$\pm$0.14 & 16.3$\pm$1.8 \\
BEVFusion~\cite{liu2023bevfusion} & 1 & C & 8.8$\pm$1.9 & 13.6$\pm$5.2 & 0.77$\pm$0.03 & 1.29$\pm$0.10 & 17.4$\pm$1.8 \\
\midrule
BEVFusion~\cite{liu2023bevfusion} & 5 & A (held-in) & 17.1$\pm$0.4 & 30.4$\pm$1.1 & 0.66$\pm$0.02 & 0.87$\pm$0.02 & 25.0$\pm$0.2 \\
BEVFusion~\cite{liu2023bevfusion} & 5 & B & 15.1$\pm$0.9 & 31.1$\pm$4.0 & 0.60$\pm$0.01 & 1.18$\pm$0.03 & 21.5$\pm$0.6 \\
BEVFusion~\cite{liu2023bevfusion} & 5 & C & 23.7$\pm$1.1 & 44.0$\pm$1.7 & 0.66$\pm$0.01 & 0.95$\pm$0.02 & 27.5$\pm$0.7 \\
BEVFusion~\cite{liu2023bevfusion} & 5 & D & 21.1$\pm$1.8 & 25.2$\pm$5.7 & 0.65$\pm$0.01 & 0.97$\pm$0.05 & 25.0$\pm$1.1 \\
BEVFusion~\cite{liu2023bevfusion} & 5 & E & 26.7$\pm$1.1 & 39.9$\pm$2.5 & 0.71$\pm$0.00 & 1.30$\pm$0.03 & 28.0$\pm$0.6 \\
\bottomrule
\end{tabular}
}{\tbd{run the harness}}
\end{table*}
\begin{table*}[t]
  \centering\footnotesize
  \caption{Per-class AP in percent on the earlier fused-cloud scenes, mean over three seeds. \srcnote}
  \label{stab:perclass}
  \IfFileExists{table_perclass.tex}{\begin{tabular}{@{}lrrrrrrrrr@{}}
\toprule
Class & \multicolumn{3}{c}{PointPillars} & \multicolumn{3}{c}{CenterPoint} & \multicolumn{3}{c}{BEVFusion} \\
\cmidrule(lr){2-4}\cmidrule(lr){5-7}\cmidrule(lr){8-10}
 & A & B & C & A & B & C & A & B & C \\
\midrule
Car & 23.4 & 0.5 & 5.4 & 9.6 & 0.0 & 3.0 & 27.7 & 4.4 & 7.2 \\
Truck & 4.4 & 0.0 & 0.1 & 3.2 & 0.0 & 0.0 & 4.4 & 0.0 & 0.3 \\
Bus & 1.8 & 0.0 & 0.0 & 1.9 & 0.0 & 0.0 & 0.8 & 0.0 & 0.0 \\
Motorcycle & 0.5 & 0.0 & 0.0 & 7.3 & 0.0 & 2.4 & 28.5 & 10.6 & 19.8 \\
Bicycle & 0.0 & 0.0 & 0.0 & 2.9 & 0.0 & 0.0 & 1.7 & 0.7 & 0.5 \\
Pedestrian & 31.4 & 1.1 & 4.4 & 30.1 & 0.4 & 7.1 & 47.3 & 3.2 & 15.6 \\
CNG auto-rickshaw & 17.9 & 0.0 & 1.7 & 14.9 & 0.2 & 3.9 & 25.8 & 23.6 & 18.1 \\
Cycle rickshaw & 30.1 & 0.0 & 0.8 & 16.8 & 0.0 & 0.1 & 37.8 & 0.8 & 9.0 \\
\bottomrule
\end{tabular}
}{\tbd{run the harness}}
\end{table*}
\subsection{3D Object Detection}
Tables~\ref{stab:detfull} to~\ref{stab:perclass} give the full metric set on the held-in split, the per-sequence results of every model including the five-sequence BEVFusion, and the per-class breakdown.
Orientation error exceeds one radian for every model on the earlier fused-cloud scenes, and the scale error of about 0.55 reflects the surface-fitted boxes rather than the model.

\textbf{Diagnostics for the Density Reading:}
Four measurements on the single-sequence models identify scene density as the variable that governs held-out accuracy.
\begin{itemize}\setlength{\itemsep}{2pt}
  \item \textbf{The models localise.} A same-class prediction lies within \SI{4}{\metre} of \LocalisedPedEightPct\,\% of pedestrians on sequence B and \LocalisedPedNinePct\,\% on sequence C, against 90\,\% on the held-in split.
  \item \textbf{Accuracy is not limited by localisation.} On sequence B the AP of CenterPoint is flat across every match radius, at 0.008 for \SI{0.5}{\metre} and for \SI{4}{\metre}, where on the held-in split it climbs from 0.178 to 0.489.
  \item \textbf{Scores carry little ranking information.} Only \ScoreAboveHalfPct\,\% of predictions exceed 0.5, and the probability that a true positive outranks a false positive is 0.64 for PointPillars pedestrians on the held-in split and 0.60 on sequence B, where 0.5 is chance.
  \item \textbf{The number of proposals is constant.} CenterPoint emits about 490, 398 and 468 boxes per keyframe on sequences A, B and C and PointPillars 143, 83 and 123, while the ground truth falls from \GtPerFrameZero{} to \GtPerFrameEight{} and \GtPerFrameNine{} objects per keyframe.
\end{itemize}
Precision is therefore governed by how often a fixed number of proposals coincides with a real object, and average precision on the earlier fused-cloud scenes tracks scene density.
The five-sequence BEVFusion of Table~\ref{stab:gen} narrows the ratio between its best and worst sequence from \GenSpreadBEVFusionOne$\times$ to \GenSpreadBEVFusionFive$\times$ with a seed spread of at most 2.2 points, and reaches \PooledCngHeldOutAP\,\% AP on the CNG auto-rickshaw of sequence B.

\begin{table}[t]
\caption{Input point cloud ablation. CenterPoint on the Mid-360 returns alone and on the fused LiDAR and stereo-depth cloud of the earlier fused-cloud scenes, pooled split over sequences A to C. mAP per sequence with standard deviation over three seeds, and mean mAP and NDS, in percent. \srcnote}
\label{stab:ablation}
\centering\scriptsize
\IfFileExists{table_ablation.tex}{\begin{adjustbox}{max width=\columnwidth}\begin{tabular}{@{}lrrrrr@{}}
\toprule
& \multicolumn{3}{c}{mAP by sequence} & \multicolumn{2}{c}{Mean} \\
\cmidrule(lr){2-4}\cmidrule(lr){5-6}
Input cloud & A & B & C & mAP & NDS \\
\midrule
Mid-360 only & \textbf{6.8$\pm$0.2} & \textbf{9.5$\pm$0.3} & \textbf{14.7$\pm$0.6} & \textbf{10.3} & \textbf{18.4} \\
Mid-360 + stereo depth & 4.8$\pm$0.1 & 0.6$\pm$0.1 & 5.5$\pm$0.3 & 3.7 & 12.7 \\
\bottomrule
\end{tabular}
\end{adjustbox}}{\tbd{run the harness}}
\end{table}

\textbf{Effect of the Input Point Cloud:}
Table~\ref{stab:ablation} trains CenterPoint identically except for the input.
The Mid-360 rings alone, about \SI{40}{k} points per keyframe against \SI{414}{k} for the fused cloud, win on every sequence by more than the seed spread, and KISS-ICP on the same release reaches \KissDayOneMidATE\,\si{\metre} ATE on the Mid-360 points against \KissDayOneFusedATE\,\si{\metre} on the fused cloud and \KissDayOneStereoATE\,\si{\metre} on the stereo points alone.
The rear stereo camera's depth cloud is tilted \SI{4}{\degree} against the LiDAR plane, so every object acquires a displaced copy, and the fused arm additionally reached the voxel cap of the detector on every keyframe.
The four benchmarked scenes separates the clouds. By oriented containment a box holds a median of 292 to 397 stereo points against \MidPtsPerBoxLo{} to \MidPtsPerBoxHi{} Mid-360 points, and \MidEmptyBoxLoPct{} to \MidEmptyBoxHiPct\,\% of boxes contain no Mid-360 point, so LiDAR-only detectors on that release are evaluated against boxes that are partly observable only in stereo.
The stereo channel is stored in the world frame with an identity sensor and ego pose. It is read into the ego frame through the inverse of the LiDAR channel's ego pose, which puts cloud and boxes in one frame, and the benchmarked-scene detectors use it that way.

\begin{table*}[t]
\caption{3D detection and point segmentation on the four benchmarked scenes per sequence (F to I), mean and standard deviation over three seeds. Percent except mATE in m and mAOE in rad. \srcnote}
\label{stab:daytwo}
\centering\scriptsize
\IfFileExists{table_daytwo_full.tex}{\begin{tabular}{@{}llrrrrrr@{}}
\toprule
Method & Seq. & mAP & Rickshaw AP & mATE & mAOE & NDS & mIoU \\
\midrule
CenterPoint~\cite{yin2021centerpoint} & F & 37.5$\pm$0.7 & 55.0$\pm$1.7 & 0.42$\pm$0.02 & 1.41$\pm$0.05 & 35.9$\pm$0.4 & 20.1$\pm$0.1 \\
CenterPoint~\cite{yin2021centerpoint} & G & 33.1$\pm$0.3 & 38.7$\pm$0.7 & 0.45$\pm$0.01 & 1.37$\pm$0.09 & 33.5$\pm$0.4 & 14.2$\pm$0.2 \\
CenterPoint~\cite{yin2021centerpoint} & H & 38.2$\pm$0.6 & 61.2$\pm$1.9 & 0.44$\pm$0.01 & 1.26$\pm$0.05 & 35.9$\pm$0.4 & 17.9$\pm$0.2 \\
CenterPoint~\cite{yin2021centerpoint} & I & 46.1$\pm$0.3 & 61.8$\pm$1.3 & 0.41$\pm$0.00 & 1.53$\pm$0.02 & 41.5$\pm$0.3 & 18.7$\pm$0.6 \\
\midrule
BEVFusion~\cite{liu2023bevfusion} & F & 29.5$\pm$6.9 & 51.6$\pm$15.8 & 0.53$\pm$0.03 & 1.30$\pm$0.08 & 28.7$\pm$3.1 & 10.7$\pm$1.6 \\
BEVFusion~\cite{liu2023bevfusion} & G & 28.6$\pm$7.5 & 36.7$\pm$17.8 & 0.51$\pm$0.05 & 1.22$\pm$0.07 & 27.4$\pm$3.7 & 8.4$\pm$0.5 \\
BEVFusion~\cite{liu2023bevfusion} & H & 31.0$\pm$5.7 & 51.8$\pm$10.2 & 0.60$\pm$0.02 & 1.32$\pm$0.01 & 28.4$\pm$2.4 & 9.9$\pm$0.6 \\
BEVFusion~\cite{liu2023bevfusion} & I & 36.1$\pm$7.0 & 53.7$\pm$16.0 & 0.48$\pm$0.07 & 1.37$\pm$0.09 & 33.4$\pm$4.3 & 9.4$\pm$0.2 \\
\midrule
PointPillars~\cite{lang2019pointpillars} & F & 22.9$\pm$0.1 & 36.5$\pm$0.2 & 0.63$\pm$0.00 & 1.22$\pm$0.02 & 23.4$\pm$0.0 & 13.2$\pm$0.4 \\
PointPillars~\cite{lang2019pointpillars} & G & 19.1$\pm$0.5 & 29.2$\pm$1.7 & 0.57$\pm$0.01 & 1.31$\pm$0.03 & 23.6$\pm$0.6 & 8.8$\pm$0.2 \\
PointPillars~\cite{lang2019pointpillars} & H & 23.1$\pm$0.2 & 41.0$\pm$0.4 & 0.56$\pm$0.04 & 1.27$\pm$0.04 & 27.3$\pm$1.2 & 10.3$\pm$0.3 \\
PointPillars~\cite{lang2019pointpillars} & I & 31.3$\pm$0.7 & 55.3$\pm$2.3 & 0.54$\pm$0.04 & 1.24$\pm$0.13 & 32.7$\pm$0.8 & 13.6$\pm$0.7 \\
\midrule
TransFusion-L~\cite{bai2022transfusion} & F & 21.1$\pm$7.0 & 24.2$\pm$11.4 & 0.53$\pm$0.02 & 1.46$\pm$0.10 & 24.5$\pm$5.4 & 7.1$\pm$0.5 \\
TransFusion-L~\cite{bai2022transfusion} & G & 19.8$\pm$6.5 & 15.0$\pm$9.1 & 0.56$\pm$0.10 & 1.40$\pm$0.06 & 23.3$\pm$5.3 & 4.8$\pm$0.6 \\
TransFusion-L~\cite{bai2022transfusion} & H & 22.7$\pm$7.9 & 32.7$\pm$19.6 & 0.58$\pm$0.06 & 1.38$\pm$0.05 & 24.7$\pm$6.0 & 6.1$\pm$0.6 \\
TransFusion-L~\cite{bai2022transfusion} & I & 23.3$\pm$9.4 & 22.7$\pm$14.5 & 0.55$\pm$0.10 & 1.35$\pm$0.19 & 26.4$\pm$7.9 & 7.0$\pm$0.3 \\
\bottomrule
\end{tabular}
}{\tbd{run the harness}}
\end{table*}
\begin{table}[t]
\caption{Per-class AP in percent on the four benchmarked scenes, mean over three seeds and four sequences. \srcnote}
\label{stab:daytwopc}
\centering\scriptsize
\IfFileExists{table_daytwo_perclass.tex}{\begin{adjustbox}{max width=\columnwidth}\begin{tabular}{@{}lrrrr@{}}
\toprule
Class & CenterPoint & BEVFusion & PointPillars & TransFusion-L \\
\midrule
Car & 49.1 & 29.7 & 45.4 & 40.8 \\
Truck & 17.3 & 13.0 & 13.6 & 9.4 \\
Bus & 1.0 & 1.5 & 1.0 & 0.9 \\
Motorcycle & 70.5 & 50.6 & 0.4 & 22.5 \\
Bicycle & 0.3 & 0.1 & 0.0 & 0.0 \\
Pedestrian & 63.4 & 58.5 & 51.1 & 53.1 \\
CNG auto-rickshaw & 35.7 & 35.5 & 19.9 & 14.3 \\
Cycle rickshaw & 72.6 & 61.4 & 61.1 & 33.0 \\
\bottomrule
\end{tabular}
\end{adjustbox}}{\tbd{run the harness}}
\end{table}
\begin{table}[t]
\caption{Per-class IoU in percent of the box-derived point segmentation on the stereo layer of the four benchmarked scenes, mean over three seeds and four sequences. Best per class in bold. \srcnote}
\label{stab:segpc}
\centering\scriptsize
\IfFileExists{table_seg_perclass.tex}{\begin{adjustbox}{max width=\columnwidth}\begin{tabular}{@{}lrrrr@{}}
\toprule
Class & CenterPoint & BEVFusion & PointPillars & TransFusion-L \\
\midrule
Car & 33.1 & 16.1 & \textbf{30.9} & 19.3 \\
Truck & 11.0 & 9.6 & 8.1 & 2.0 \\
Bus & 0.7 & 1.4 & 0.2 & 0.0 \\
Motorcycle & 20.8 & 4.7 & 0.4 & 2.2 \\
Bicycle & 0.9 & 0.0 & 0.1 & 0.0 \\
Pedestrian & \textbf{33.9} & \textbf{37.0} & 29.2 & \textbf{25.4} \\
CNG auto-rickshaw & 18.1 & 3.5 & 7.3 & 0.2 \\
Cycle rickshaw & 23.3 & 4.4 & 15.6 & 1.1 \\
\bottomrule
\end{tabular}
\end{adjustbox}}{\tbd{run the harness}}
\end{table}

\textbf{Second Release, Detection and Segmentation:}
Tables~\ref{stab:daytwo} to~\ref{stab:segpc} report four detectors trained on the stereo cloud of sequences F to I and the point segmentation derived from their boxes.
CenterPoint reaches \DayTwoCenterPointMeanmAP\,\% mAP with a seed spread below one point, BEVFusion \DayTwoBEVFusionMeanmAP\,\% at half the epochs, PointPillars \DayTwoPointPillarsMeanmAP\,\% and TransFusion-L \DayTwoTransFusionMeanmAP\,\% with a large seed variance.
The class-prior labelling regime conditions detection far better than the fitted extents of the earlier fused-cloud scenes. A single anchor set covers \DayTwoAnchorCoveragePct\,\% of the ground-truth boxes at the strictest matching threshold, against \BoxAnchorCoveragePct\,\% on the earlier fused-cloud scenes, and the median dimensions are physically right, with a pedestrian \SI{1.72}{\metre} tall and a bus \SI{11.19}{\metre} long. Because seven of the eight classes carry one length value, mASE measures prior recovery rather than size estimation and is not reported for this release.
Segmentation is scored on the stereo layer that the detectors consume. Cameras buy pedestrians and nothing else. BEVFusion is the best model on pedestrians and worse than both LiDAR-only models on every other class, car and cycle rickshaw included. PointPillars never proposes a motorcycle, so its motorcycle AP and IoU are both near zero on every seed and sequence, one architectural cause visible in two tasks. Bus and bicycle sit near zero for every model on 1 to 53 ground-truth instances per sequence.
Confidence scales differ by two orders of magnitude across architectures, so the segmentation operating point is swept per model on one sequence and applied unchanged elsewhere, at 0.1 for the OpenPCDet LiDAR models and \SegBevfusionThreshold{} for BEVFusion. At a common threshold BEVFusion would keep two boxes out of a hundred thousand and read as zero.
The gate retained boxes to \DayTwoMaxRangeM\,\si{\metre}, so the \SI{50}{\metre} range band of the benchmark is empty on this release, and keyframes are at \SI{5}{\hertz} rather than the \SI{2}{\hertz} of nuScenes, so velocity metrics are not directly comparable.

\begin{table*}[t]
\caption{Strata on the four benchmarked scenes, NDS in percent, mean over four sequences and three seeds. Left, density bins with the change $\Delta^{\uparrow}_\rho$ from Low to Extreme in points, raw and matched over the classes present in both bins, and the composite RS with its worst density-by-illumination cell. Centre, illumination bins over the runs whose sequence contains the bin (the dark bin is empty on this release). Right, range bands (the band beyond \SI{40}{\metre} is empty). A dagger marks a stratum flagged as thinly supported in at least one run. \srcnote}
\label{stab:strata}
\centering\scriptsize
\IfFileExists{table_strata.tex}{\begin{tabular}{@{}lrrrrrr@{}}
\toprule
& \multicolumn{4}{c}{NDS by density bin} & \multicolumn{2}{c}{$\Delta^{\uparrow}_\rho$} \\
\cmidrule(lr){2-5}\cmidrule(lr){6-7}
Method & Low & Medium & High & Extreme & Raw & Matched \\
\midrule
CenterPoint & 30.2$^{\dagger}$ & 36.2 & 35.6 & 36.5 & +6.3 & +4.1 \\
BEVFusion & 23.4$^{\dagger}$ & 29.2 & 29.3 & 29.8 & +6.4 & +6.3 \\
PointPillars & 21.9$^{\dagger}$ & 25.8 & 26.1 & 26.0 & +4.1 & +1.8 \\
TransFusion-L & 19.7$^{\dagger}$ & 23.4 & 24.5 & 26.0 & +6.3 & +5.6 \\
\bottomrule
\end{tabular}
}{\tbd{run the harness}}
\hspace{1.5em}
\IfFileExists{table_illum.tex}{\begin{tabular}{@{}lrrr@{}}
\toprule
& \multicolumn{3}{c}{NDS by illumination bin} \\
\cmidrule(lr){2-4}
Method & Day & Dusk & Night \\
\midrule
CenterPoint & 35.2 & 35.3 & 37.2$^{\dagger}$ \\
BEVFusion & 30.0 & 28.8 & 27.8$^{\dagger}$ \\
PointPillars & 23.5 & 26.1 & 25.0$^{\dagger}$ \\
TransFusion-L & 25.2 & 24.1 & 23.7$^{\dagger}$ \\
\bottomrule
\end{tabular}
}{\tbd{run the harness}}
\hspace{1.5em}
\IfFileExists{table_range.tex}{\begin{tabular}{@{}lrr@{}}
\toprule
& \multicolumn{2}{c}{NDS by range band} \\
\cmidrule(lr){2-3}
Method & 0--20 m & 20--40 m \\
\midrule
CenterPoint & 36.8 & 27.5 \\
BEVFusion & 30.1 & 23.7 \\
PointPillars & 26.3 & 23.1 \\
TransFusion-L & 25.2 & 20.3 \\
\bottomrule
\end{tabular}
}{\tbd{run the harness}}
\end{table*}
\begin{table}[t]
\caption{Density strata on the earlier fused-cloud scenes per evaluation sequence (A held-in, B and C held-out), NDS in percent, mean over three seeds, with $\Delta^{\uparrow}_\rho$ from Low to Extreme. A dagger marks a stratum with too few ground-truth boxes to be reliable. \srcnote}
\label{stab:stratadayone}
\centering\scriptsize
\IfFileExists{table_strata_dayone.tex}{\begin{adjustbox}{max width=\columnwidth}\begin{tabular}{@{}llrrrrrr@{}}
\toprule
& & \multicolumn{4}{c}{NDS by density bin} & \multicolumn{2}{c}{$\Delta^{\uparrow}_\rho$} \\
\cmidrule(lr){3-6}\cmidrule(lr){7-8}
Method & Seq. & Low & Medium & High & Extreme & Raw & Matched \\
\midrule
BEVFusion & A (held-in) & 26.0 & 26.9 & 26.0 & 23.6 & -2.4 & -2.4 \\
BEVFusion & B & 2.5$^{\dagger}$ & 13.4 & 17.8 & 15.6 & +13.1 & +7.1 \\
BEVFusion & C & 12.7 & 16.7 & 16.6 & 18.6 & +5.9 & +4.5 \\
\midrule
CenterPoint & A (held-in) & 19.7 & 17.5 & 19.3 & 17.6 & -2.1 & -2.1 \\
CenterPoint & B & 0.0$^{\dagger}$ & 6.1 & 6.2 & 7.0 & +7.0 & +4.2 \\
CenterPoint & C & 5.1 & 7.6 & 9.2 & 11.2 & +6.1 & +5.9 \\
\midrule
PointPillars & A (held-in) & 23.1 & 22.2 & 21.7 & 19.4 & -3.7 & -3.7 \\
PointPillars & B & 0.0$^{\dagger}$ & 5.0 & 6.0 & 6.2 & +6.2 & +6.5 \\
PointPillars & C & 4.9 & 7.5 & 9.4 & 11.3 & +6.4 & +6.2 \\
\bottomrule
\end{tabular}
\end{adjustbox}}{\tbd{run the harness}}
\end{table}
\subsection{Stratified Evaluation}
Table~\ref{stab:strata} gives the density, illumination and range strata of every benchmarked-scene model.
Accuracy rises with crowding for all four detectors, by \DensDeltaCenterPoint{} NDS points from Low to Extreme for CenterPoint and \DensDeltaPointPillars{} for PointPillars, in both the raw and the matched-class form.
Density on this route is partly a proxy for range in the direction opposite to the intuition. A low-density keyframe is an open stretch whose few objects are distant, and distance is what hurts, since the per-class effective range of the labels is 25 to \SI{31}{\metre}. The range bands make it explicit, with every model losing several NDS points from the 0 to \SI{20}{\metre} band to the 20 to \SI{40}{\metre} band. The composite confirms it, with the worst cell for every detector at Low density and night, never a crowded cell, and composites of \RsCenterPoint{} for CenterPoint and \RsPointPillars{} for PointPillars.
The Low bin carries \StrataGtLow{} ground-truth objects against \StrataGtExtreme{} in Extreme, so it is the noisiest estimate, and the dark and 40 m+ strata contain no ground truth, so the composite is computed over a partial cell grid.
Illumination is read on this release without the route confound of the first, and within the noise the three bins are level for CenterPoint and slightly lower at night for the others.
Table~\ref{stab:stratadayone} keeps the first-release strata. On the held-in sequence every detector scored a few points lower at Extreme than at Low, and on the held-out sequences the sparsest bin was worst by a wide margin, which is the constant-proposal mechanism of the held-out result observed within a sequence. The benchmarked-scene result supersedes it in coverage, with four sequences and four detectors against one sequence per model.

\begin{table*}[t]
\caption{Odometry on the four benchmarked scenes per sequence, F being the sequence with the recording gap and G to I the clean ones. $n_{\mathrm{est}}$ is the number of estimated poses and $n$ the number of \SI{1}{\hertz} reference pairs associated within \SI{50}{\milli\second} after SE(3) Umeyama alignment. ATE is the RMSE in m and degrees, RPE the translational RMSE at \SI{1}{\second}, $t_{\mathrm{rel}}$ and $r_{\mathrm{rel}}$ the KITTI relative errors over 100 to \SI{800}{\metre} sub-sequences for converged runs, and the last column the ATE of the same estimator with the PPK solution fused as a position input, or diverged for an ATE above \SI{100}{\metre}. \srcnote}
\label{stab:odom}
\centering\scriptsize
\IfFileExists{table_odom_full.tex}{\begin{tabular}{@{}lllrrrrrrrr@{}}
\toprule
Method & Sensors & Seq. & $n_{\mathrm{est}}$ & $n$ & ATE (m) & ATE ($^\circ$) & RPE$_{1\,\mathrm{s}}$ (m) & $t_{\mathrm{rel}}$ (\%) & $r_{\mathrm{rel}}$ & ATE$_{+\mathrm{PPK}}$ (m) \\
\midrule
KISS-ICP~\cite{vizzo2023kissicp} & L & F & 1291 & 210 & 4.93 & 50.9 & 5.49 & 54.4 & 0.203 & 1.41 \\
KISS-ICP~\cite{vizzo2023kissicp} & L & G & 1490 & 259 & 6.00 & 33.2 & 3.49 & 44.3 & 0.164 & 1.50 \\
KISS-ICP~\cite{vizzo2023kissicp} & L & H & 1491 & 296 & 2.93 & 15.7 & 1.33 & 11.7 & 0.088 & 1.93 \\
KISS-ICP~\cite{vizzo2023kissicp} & L & I & 1488 & 298 & 5.71 & 42.3 & 1.41 & 11.0 & 0.035 & 2.62 \\
\midrule
GLIM~\cite{koide2024glim} & L+I & F & 1283 & 210 & 3.57 & 53.3 & 5.54 & 60.5 & 0.201 & 1.38 \\
GLIM~\cite{koide2024glim} & L+I & G & 1474 & 256 & 4.03 & 33.4 & 3.50 & 44.4 & 0.165 & 1.53 \\
GLIM~\cite{koide2024glim} & L+I & H & 1475 & 293 & 2.33 & 15.3 & 1.33 & 11.2 & 0.085 & 1.84 \\
GLIM~\cite{koide2024glim} & L+I & I & 1473 & 295 & 3.28 & 44.4 & 1.39 & 11.1 & 0.032 & 3.15 \\
\midrule
LIO-SAM~\cite{shan2020liosam} & L+I & F & 519 & 78 & 603.43 & 133.8 & 137.64 & -- & -- & Diverged \\
LIO-SAM~\cite{shan2020liosam} & L+I & G & 530 & 98 & 5.08 & 30.8 & 8.92 & 41.5 & 0.154 & 1.43 \\
LIO-SAM~\cite{shan2020liosam} & L+I & H & 460 & 95 & 2.62 & 18.1 & 4.13 & 13.7 & 0.109 & 2.10 \\
LIO-SAM~\cite{shan2020liosam} & L+I & I & 480 & 98 & 634.29 & 122.0 & 114.16 & -- & -- & Diverged \\
\midrule
FAST-LIVO2~\cite{zheng2025fastlivo2} & L+I+C & F & 1288 & 210 & 92647.57 & 124.1 & 2390.71 & -- & -- & Diverged \\
FAST-LIVO2~\cite{zheng2025fastlivo2} & L+I+C & G & 1487 & 258 & 4.10 & 33.5 & 3.50 & 44.9 & 0.164 & 1.55 \\
FAST-LIVO2~\cite{zheng2025fastlivo2} & L+I+C & H & 1488 & 295 & 2.77 & 15.3 & 1.40 & 11.3 & 0.086 & 1.80 \\
FAST-LIVO2~\cite{zheng2025fastlivo2} & L+I+C & I & 1485 & 297 & 205.35 & 171.1 & 7.90 & -- & -- & Diverged \\
\midrule
ZED VIO (front) & S+I & F & 7787 & 252 & 18.18 & 48.1 & 7.54 & 52.1 & 0.205 & 6.95 \\
ZED VIO (front) & S+I & G & 8946 & 299 & 15.51 & 36.3 & 2.96 & 49.1 & 0.163 & 1.44 \\
ZED VIO (front) & S+I & H & 8964 & 298 & 7.84 & 15.7 & 1.36 & 13.5 & 0.092 & 1.87 \\
ZED VIO (front) & S+I & I & 8984 & 301 & 45.18 & 51.3 & 1.55 & 26.9 & 0.056 & 2.91 \\
\midrule
ZED VIO (rear) & S+I & F & 7790 & 253 & 8.71 & 48.0 & 7.55 & 51.5 & 0.202 & 6.76 \\
ZED VIO (rear) & S+I & G & 8939 & 299 & 19.93 & 36.5 & 3.03 & 51.2 & 0.162 & 1.67 \\
ZED VIO (rear) & S+I & H & 8978 & 299 & 10.75 & 15.4 & 1.40 & 15.6 & 0.092 & 2.05 \\
ZED VIO (rear) & S+I & I & 8967 & 300 & 40.84 & 61.2 & 1.69 & 23.1 & 0.039 & 3.53 \\
\midrule
ORB-SLAM3~\cite{campos2021orbslam3} (front) & S+I & F & 18 & 2 & 0.02 & 86.8 & 3.96 & -- & -- & Diverged \\
ORB-SLAM3~\cite{campos2021orbslam3} (front) & S+I & G & 1129 & 107 & 8.43 & 138.9 & 9.15 & 131.1 & 0.162 & 4.07 \\
ORB-SLAM3~\cite{campos2021orbslam3} (front) & S+I & H & 455 & 45 & 8.57 & 89.2 & 5.89 & 124.7 & 0.038 & 2.82 \\
ORB-SLAM3~\cite{campos2021orbslam3} (front) & S+I & I & 458 & 40 & 4.47 & 176.3 & 11.96 & 132.0 & 0.101 & 3.50 \\
\midrule
ORB-SLAM3~\cite{campos2021orbslam3} (rear) & S+I & F & 0 & 0 & -- & -- & -- & -- & -- & Diverged \\
ORB-SLAM3~\cite{campos2021orbslam3} (rear) & S+I & G & -- & -- & -- & -- & -- & -- & -- & -- \\
ORB-SLAM3~\cite{campos2021orbslam3} (rear) & S+I & H & -- & -- & -- & -- & -- & -- & -- & -- \\
ORB-SLAM3~\cite{campos2021orbslam3} (rear) & S+I & I & -- & -- & -- & -- & -- & -- & -- & -- \\
\bottomrule
\end{tabular}
}{\tbd{run the harness}}
\end{table*}
\subsection{Odometry}
\label{supp:odom}
\textbf{Ground Truth:}
The PPK solution of the four benchmarked scenes has \PpkDayTwoEpochs{} epochs at \SI{1}{\hertz}, \PpkDayTwoFixedPct\,\% fixed at a horizontal sigma of \PpkSigmaFixedMm\,\si{\milli\metre} and \PpkDayTwoFloatPct\,\% float at \PpkSigmaFloatMm\,\si{\milli\metre}.
The reference trajectory is built from the raw solution. The solution is stamped in GPS time and its offset to the receiver's UTC-stamped fixes was measured at \PpkGpstOffsetS\,\si{\second} with a residual of \SI{1.59}{\metre}. Heading comes from the dual-antenna stream, whose baseline is mounted across the vehicle, so a boresight offset of \PpkBoresightDeg\,\si{\degree} is applied as a mounting constant. It scatters by \PpkBoresightMadDeg\,\si{\degree} median absolute deviation over the moving epochs, and after removing it the heading agrees with the trajectory's own course to \PpkHeadingAgreeDeg\,\si{\degree}.

\textbf{Preparing the Export for SLAM:}
Three inputs that LiDAR-inertial SLAM expects are reconstructed from the nuScenes-format export. Per-point time is recovered from the firing order, since consecutive points differ in range by a median of \MidRangeStepCm\,\si{\centi\metre} against \SI{4.6}{\metre} when re-sorted by azimuth. IMU orientation is derived from the IMU itself with a Mahony filter that tracks smoothed gravity to within \SI{0.7}{\degree} on all four sequences. Clouds are written one sweep late in bag time so that the IMU leads the cloud, with headers unchanged. Estimators that stamp a pose at their update instant rather than at the sweep are snapped to the sweep they describe before association, since a \SI{15}{\milli\second} offset inside the \SI{50}{\milli\second} window otherwise discards most reference pairs and biases the survivors.

\textbf{Results:}
Table~\ref{stab:odom} reports every estimator per sequence, with and without the PPK solution fused as a position input. Rotation is not fused because one antenna does not observe attitude.
On the clean sequences GLIM and KISS-ICP hold track throughout with \OdomGLIMMinATE{} to \OdomGLIMMaxATE{} and \OdomKISSICPMinATE{} to \OdomKISSICPMaxATE\,\si{\metre} ATE, and the tightly coupled LiDAR-inertial systems are competitive where they hold track. FAST-LIVO2 gives the best LiDAR-inertial ATE on sequence G and LIO-SAM beats KISS-ICP on sequence H, though LIO-SAM emits a pose for only about 40\,\% of sweeps in either mode.
Two perturbations in the data separate the groups. Sequence F is missing \DropoutLidarS\,\si{\second} of LiDAR from \SI{1.4}{\second} in, with a \DropoutImuS\,\si{\second} IMU gap and a PPK gap nested inside it, a recording dropout rather than a sensor fault. Sequence I carries the largest shock in the release, \ShockPeakAccel\,\si{\metre\per\second\squared} peak acceleration (\ShockPeakG{}\,g), and the largest single jump of LIO-SAM lands about six seconds after it. LIO-SAM and FAST-LIVO2 diverge on both sequences by two orders of magnitude, FAST-LIVO2 through accumulated drift after the shock rather than one discontinuity, while GLIM, KISS-ICP and ORB-SLAM3 handle both, ORB-SLAM3 scoring its best sequence on the one with the shock. Compute starvation, point ordering, a cloud-to-IMU frame mismatch and raw IMU quality were each tested and excluded as causes.
The stereo visual-inertial estimators trail the LiDAR methods by a factor of five to seven without GNSS and match them once the PPK solution is fused. Every method's yaw disagrees with its own course over ground by tens of degrees, so the rotation columns are listed and not interpreted.

\begin{table}[t]
\caption{Stereo depth of the front ZED against the LiDAR projection, per sequence (F to I) and per illumination bin averaged over sequences. \srcnote}
\label{stab:depth}
\centering\scriptsize
\IfFileExists{table_depth.tex}{\begin{adjustbox}{max width=\columnwidth}\begin{tabular}{@{}lrrrrr@{}}
\toprule
Seq.\,/\,bin & Frames & AbsRel & RMSE (m) & $\delta<1.25$ & $\delta<1.25^{3}$ \\
\midrule
F & 251 & 0.839 & 9.20 & 0.281 & 0.707 \\
G & 299 & 0.869 & 10.28 & 0.278 & 0.665 \\
H & 299 & 0.922 & 10.36 & 0.333 & 0.687 \\
I & 299 & 1.048 & 11.92 & 0.258 & 0.603 \\
\midrule
Mean &  & 0.919 & 10.44 & 0.287 & 0.665 \\
\midrule
Day & 350 & 0.904 & 10.37 & 0.283 & 0.665 \\
Dusk & 585 & 0.884 & 10.19 & 0.291 & 0.679 \\
Night & 213 & 1.081 & 11.75 & 0.283 & 0.610 \\
\bottomrule
\end{tabular}
\end{adjustbox}}{\tbd{run the harness}}
\end{table}
\begin{table}[t]
\caption{Localization of the PPK-fused estimators, RMSE of the along-track and cross-track error in m over the sequences run, their ratio, translation-only success within \SI{0.5}{\metre}, and the worst sequence. \srcnote}
\label{stab:loc}
\centering\scriptsize
\IfFileExists{table_loc.tex}{\begin{adjustbox}{max width=\columnwidth}\begin{tabular}{@{}lrrrrrr@{}}
\toprule
Method & Seq. & Longitudinal (m) & Lateral (m) & Ratio & Success$_{0.5\,\mathrm{m}}$ (\%) & Worst \\
\midrule
GLIM + PPK & 4 & 1.37 & 1.07 & 1.28 & 50 & I \\
KISS-ICP + PPK & 4 & 1.38 & 0.83 & 1.66 & 47 & I \\
LIO-SAM + PPK & 4 & 33 & 43 & -- & -- & F \\
FAST-LIVO2 + PPK & 2 & 871 & 544 & -- & -- & F \\
\bottomrule
\end{tabular}
\end{adjustbox}}{\tbd{run the harness}}
\end{table}
\begin{table}[t]
\caption{Occupancy from a single sweep against the $\pm4$ keyframe accumulation, per sequence, in percent. \srcnote}
\label{stab:occ}
\centering\scriptsize
\IfFileExists{table_occ.tex}{\begin{adjustbox}{max width=\columnwidth}\begin{tabular}{@{}lrrr@{}}
\toprule
Seq. & Frames & Voxel mIoU & RayIoU \\
\midrule
F & 251 & 17.0 & 74.1 \\
G & 299 & 18.5 & 68.6 \\
H & 299 & 22.5 & 71.4 \\
I & 299 & 17.3 & 78.2 \\
\midrule
Mean &  & 18.8 & 73.1 \\
\bottomrule
\end{tabular}
\end{adjustbox}}{\tbd{run the harness}}
\end{table}
\subsection{Depth, Localization and Occupancy}
None of these tasks has a trained baseline, so the rig's own sensors are scored, which characterizes the data rather than a model.

\textbf{Depth:} Table~\ref{stab:depth} scores the front ZED's reprojected cloud against the LiDAR projection. Scored raw, the cloud covers about a tenth of the image and the LiDAR projection about \DepthGtCoveragePct\,\% of it, and the two coincide on about \DepthCoincidentPx{} pixels per frame, so most reference pixels would meet no prediction and the score would measure the sampling pattern. With the 8-pixel fill the stereo reaches \DepthAbsRelMean{} AbsRel and \DepthRmseMean\,\si{\metre} RMSE over the four sequences. Night costs \DepthNightAbsRelPenaltyPct\,\% in AbsRel and \DepthNightRmsePenaltyPct\,\% in RMSE against day, the LiDAR reference being unaffected by construction. The ZED SDK's dense depth would be the fairer entry and is not yet scored.

\textbf{Localization:} Table~\ref{stab:loc} splits the PPK-fused error. GLIM and KISS-ICP reach \LocGLIMLat{} and \LocKISSICPLat\,\si{\metre} laterally with along-track error \LocGLIMRatio{} to \LocKISSICPRatio{} times larger and biases below \SI{0.25}{\metre}. LIO-SAM and FAST-LIVO2 carry their odometry divergences on the recording-gap and shock sequences into localization, on the same sequences and for the same reasons, which is a consistency check passing rather than a new failure. The joint success rate at \SI{0.5}{\metre} and \SI{2}{\degree} is zero for every arm because the fusion does not correct orientation, so the translation-only rate is reported and the joint number is not relaxed.

\textbf{Occupancy:} Table~\ref{stab:occ} gives voxel mIoU of \OccVoxelMean\,\% against RayIoU of \OccRayMean\,\%, a factor of \OccRatio{} that holds on every sequence. A single \SI{5}{\hertz} sweep fills \OccSweepFillPct\,\% of the grid against \OccAccumFillPct\,\% for the accumulation, so the voxel score is limited by how little one sweep sees, while its first-hit surfaces are right about three times in four. Reporting voxel mIoU alone would have framed a sampling gap as an accuracy gap.

\subsection{Tasks Pending}
The road layer of the point segmentation is released and scorable but has no baseline, because every detection-derived method is ineligible for a stuff class and a road segmenter has not been run. Camera-only 3D detectors await a framework build for the Blackwell GPU. Tracking is withheld until a delivery with persistent instance identities.

\section{Status and Roadmap}
\label{supp:roadmap}
The following items are known to be incomplete at submission.
\begin{itemize}\setlength{\itemsep}{2pt}
  \item \textbf{Human pass.} CVAT tasks for the released scenes are published but no annotator has completed one, so every released box is pipeline output and the agreement statistics of the human boundary section are pending. Sixty-five keyframes of the second session are selected for double annotation and the second pass is not imported.
  \item \textbf{Remaining scenes.} Five of the eleven scenes of the first session are released, and four scenes of the second session are annotated and benchmarked. The rest of the second collection day is exported but not yet annotated.
  \item \textbf{Splits and strata.} No route-disjoint split exists, so every baseline uses a provisional split and there is no leaderboard. Density and illumination strata are populated on the first session with the route confound stated.
  \item \textbf{Benchmarks still to run.} A road segmenter for the LiDAR layer, the camera-only detectors, which wait on a framework build, and learned baselines for depth, localization and occupancy, where the rig's own sensors set the first mark. Tracking is withheld because the released instance identities persist for a median of \TrackMedianFrames{} keyframes.
  \item \textbf{Known label bias.} The first session's boxes are fitted to the visible surface and predate the pipeline change that drops the miscalibrated rear stereo ring, and the second session's boxes take class-prior lengths, so size errors are not comparable across releases and are not a measurement of the model on either.
  \item \textbf{Release contract.} The second session's stereo cloud is stored in the world frame with identity sensor and ego poses, and the released point count per box mixes LiDAR and stereo points, so a LiDAR-only reader must recompute it.
  \item \textbf{Calibration.} Camera latencies are unmeasured, intrinsics are nominal and the webcam and RealSense extrinsics are nominal, so boxes project cleanly only into the two stereo cameras.
  \enlargethispage{2\baselineskip}\item \textbf{Positioning.} The base station lost its L1 carrier during the first collection day, so that day's poses are code differential, and the antenna lever arm is unsurveyed. The second day's PPK reference is rebuilt from the raw solution and the export's trajectory file should not be used.
\end{itemize}

\end{document}